\documentclass[conference]{IEEEtran}
\IEEEoverridecommandlockouts    % to override locked commands

\usepackage{amsmath,amsfonts}
\usepackage[noend]{algpseudocode}
\algrenewcommand\algorithmicrequire{\textbf{Input:}}
\algrenewcommand\algorithmicensure{\textbf{Output:}}

\usepackage{algorithm}
\usepackage{array}
\usepackage[caption=false,font=normalsize,labelfont=sf,textfont=rm]{subfig}
\usepackage{textcomp}
\usepackage{stfloats}
\usepackage{url}
\usepackage{verbatim}
\usepackage{graphicx}
\usepackage{cite}
\usepackage{amssymb}
\usepackage{xcolor}
\let\labelindent\relax
\usepackage{enumitem}
\usepackage{subfig}
\usepackage{caption}
\usepackage{booktabs}
\usepackage{hyperref}
\def\BibTeX{{\rm B\kern-.05em{\sc i\kern-.025em b}\kern-.08em
    T\kern-.1667em\lower.7ex\hbox{E}\kern-.125emX}}

\title{\LARGE \bf Real-Time Generative Policy via Langevin-Guided \\ Flow Matching for Autonomous Driving}

\author{
	\parbox{\textwidth}{%
		\centering
		Tianze Zhu$^{1\dagger}$, Yinuo Wang$^{1\dagger}$, Wenjun Zou$^{1\dagger}$, Tianyi Zhang$^{1}$,
        Likun Wang$^{1}$, Letian Tao$^{1}$, Feihong Zhang$^{1}$, \\ Yao Lyu$^{1}$,
        Shengbo Eben Li$^{1,2*}$
	}%
	\thanks{$^{1}$ Tianze Zhu, Yinuo Wang, Wenjun Zou, Tianyi Zhang,
        Likun Wang, Letian Tao, Feihong Zhang and Yao Lyu are with the  School of Vehicle and Mobility, Tsinghua University, Beijing, China (E-mail: {\tt\small $\{$zhutz24, wyn23, zouwj20, zhangtia24, wlk23, tlt22, zfh24$\}$@mails.tsinghua.edu.cn}, lyo.tobias@foxmail.com)}%
	\thanks{$^{2}$ Shengbo Eben Li is with the College of Artificial Intelligence, Tsinghua University, Beijing, China (E-mail: {\tt\small lish04@gamil.com})}%
    \thanks{$^{\dagger}$ Tianze Zhu, Yinuo Wang, Wenjun Zou contribute euqually to this paper.}
    \thanks{* Corresponding Author: Shengbo Eben Li}
}

\begin{document}
	
	\maketitle
	\thispagestyle{empty}
	\pagestyle{empty}
	
	%%%%%%%%%%%%%%%%%%%%%%%%%%%%%%%%%%%%%%%%%%%%%%%%%%%%%%%%%%%%%%%%%%
	\begin{abstract}
		% Generative policies demonstrate significant promise in reinforcement learning (RL). 
        % Generative policies demonstrate significant promise in reinforcement learning (RL) for their capability to model complex distributions to enhance exploration.
        Reinforcement learning (RL) is a fundamental methodology in autonomous driving systems, where generative policies exhibit considerable potential by leveraging their ability to model complex distributions to enhance exploration. However, their inherent high inference latency severely impedes their deployment in real-time decision-making and control. To address this issue, we propose diffusion actor-critic with entropy regulator via flow matching (DACER-F) by introducing flow matching into online RL, enabling the generation of competitive actions in a single inference step. By leveraging Langevin dynamics and gradients of the Q-function, DACER-F dynamically optimizes actions from experience replay toward a target distribution that balances high Q-value information with exploratory behavior. The flow policy is then trained to efficiently learn a mapping from a simple prior distribution to this dynamic target. In complex multi-lane and intersection simulations, DACER-F outperforms baselines diffusion actor-critic with entropy regulator (DACER) and distributional soft actor-critic (DSAC), while maintaining an ultra-low inference latency. DACER-F further demonstrates its scalability on standard RL benchmark DeepMind Control Suite (DMC), achieving a score of 775.8 in the humanoid-stand task and surpassing prior methods. Collectively, these results establish DACER-F as a high-performance and computationally efficient RL algorithm.
	\end{abstract}
	
	%%%%%%%%%%%%%%%%%%%%%%%%%%%%%%%%%%%%%%%%%%%%%%%%%%%%%%%%%%%%%%%%%%
	\section{Introduction}
	\label{sec:introduction}
	Reinforcement learning (RL) is a foundational framework for sequential decision-making and control, with extensive applications in autonomous driving \cite{li2023reinforcement,lyu2025conformal,DBLP:conf/iclr/Wang0SLYCWDL25}. Conventional RL policies are typically designed to select a single optimal action. With the rising complexity and environmental uncertainty of autonomous driving, such unimodal policies often fall short of the safety and robustness requirements of advanced autonomous systems.
    Consequently, expressive policies capable of modeling complex, multimodal action distributions while enhancing exploration and generalization have become a central focus of academic research. 
    Generative models, with their powerful distribution-fitting capabilities, have emerged as the key technology for constructing such expressive policies. 
    Their remarkable success in domains such as image generation demonstrates their ability to model complex data distributions, showcasing robust representation learning capabilities.
    Within this trend, diffusion models are widely employed for constructing expressive RL policies. 
    A substantial body of research has explored the integration of diffusion models within RL, particularly in offline settings where these models have demonstrated a notable capacity for effectively modeling the distributions of expert data\cite{DBLP:conf/iclr/WangHZ23,hansen2023idql}. However, in the realm of online RL, the absence of a stationary target distribution has constrained diffusion model applications, primarily relying on complex reweighting techniques or intricate utilization of the reverse sampling process \cite{wang2024diffusion,ding2024diffusion,wang2025enhanced}. 
    
    Despite the significant promise of diffusion models for policy representation, their inherent limitation of high inference latency poses a significant barrier to autonomous driving applications. To address this critical real-time challenge, we propose to introduce flow-based generative models into autonomous driving policy learning. Flow-based models\cite{DBLP:conf/iclr/LipmanCBNL23} are a promising class of generative models, featuring efficient training and rapid inference speeds. These characteristics align perfectly with the low-latency demands of autonomous driving. Therefore, we propose the use of flow models for policy representation to significantly enhance inference efficiency. 
     However, the application of flow matching models to online RL is confronted by a core challenge: these models necessitate a well-defined target distribution $p_{\text{target}}(a|s)$, which is inherently absent in online learning settings.

    To address this limitation, we propose diffusion actor-critic with entropy regulator via flow matching (DACER-F), an algorithm that successfully integrates flow matching with online RL by introducing a dynamic target guidance mechanism. Our key contributions are as follows:
    \begin{enumerate}[label=\arabic*)]
    \item We propose a dynamic target guidance mechanism by modeling the optimal policy as an energy-based distribution induced by the Q-function, with the density given by $p(a|s) \propto \exp(Q(s,a)/\alpha)$ for $\alpha > 0$. We subsequently develop a method to efficiently sample from this energy distribution using Langevin dynamics. This process yields target action samples, denoted $a^*$, that balance high Q-value information with exploratory behavior.
    \item To the best of our knowledge, we are the first to integrate flow-matching generative models into autonomous driving policy learning under a purely RL training paradigm. By utilizing the high-quality action samples $a^*$ generated via Langevin dynamics as dynamic targets, we successfully bridge flow matching with online RL objectives, enabling the policy to efficiently learn complex mappings from a simple prior to the optimal action manifold.
    % \item Evaluation in complex multi-lane and intersection simulations demonstrates that DACER-F attains final average rewards approximately 28.0\% and 34.0\% higher than those of diffusion actor-critic with entropy regulator (DACER) \cite{wang2024diffusion} and distributional soft actor-critic (DSAC) \cite{DBLP:journals/pami/DuanWXGLLZCL25}, respectively, while reducing inference time by 84.0\% relative to DACER. The scalability of DACER-F is further validated on the DeepMind Control Suite (DMC) benchmark \cite{tassa2018deepmind}, where it consistently outperforms existing methods across six challenging tasks.
    \item We conduct extensive evaluations in complex multi-lane highway and intersection simulations, showing that DACER-F achieves final average rewards that are approximately 28.0\% and 34.0\% higher than diffusion actor-critic with entropy regulator (DACER) \cite{wang2024diffusion} and distributional soft actor-critic (DSAC) \cite{DBLP:journals/pami/DuanWXGLLZCL25}, respectively, while reducing inference time by 84.0\% relative to DACER. We further evaluate DACER-F on the DeepMind Control Suite (DMC) \cite{tassa2018deepmind}, where it consistently surpasses existing methods across six challenging tasks, demonstrating robust scalability beyond driving.
    \end{enumerate}	
    
	% Our experiments in complex multi-lane and intersection simulations show that DACER-F significantly outperforms SOTA baselines, achieving final average rewards approximately 28.0\% higher than diffusion actor-critic with entropy regulator (DACER) \cite{wang2024diffusion} and 34.0\% higher than distributional soft actor-critic (DSAC)\cite{duan2023dsac}, while concurrently achieving an 84.0\% reduction in inference time compared to DACER. Moreover, DACER-F dominates DeepMind Control Suite (DMC) \cite{tassa2018deepmind}, notably achieving 775.8 on humanoid-stand and outperforming existing algorithms by two orders of magnitude.

    %%%%%%%%%%%%%%%%%%%%%%%%%%%%%%%%%%%%%%%%%%%%%%%%%%%%%%%%%%%%%%%%%%
	\section{Related Works}
	\label{sec:conclusion}
	\subsection{Generative Policy in Offline RL}

    The use of generative models to build expressive policies is a major trend in offline RL. Among these, diffusion models have become a dominant approach, primarily leveraging their advantage in modeling complex multimodal action distributions. Pioneering studies, such as Diffusion Policy \cite{chi2023diffusion} and diffusion Q-learning (Diffusion-QL) \cite{DBLP:conf/iclr/WangHZ23}, established the use of diffusion for complex policy learning. Subsequent work integrated trajectory-level guidance, such as rewards and constraints \cite{DBLP:conf/iclr/AjayDGTJA23}.
    
    A significant line of research addresses the inherent inference latency of these diffusion-based models.
    Methods like Efficient Diffusion Policies (EDP) \cite{kang2023efficient} and consistency-based distillation \cite{DBLP:conf/atal/ChenLZ24, duan2025accelerating} approximate action generation through one-step sampling, significantly reducing computational overhead.    
    Current work continues to advance robustness and generalization, for instance, by mitigating Q-value overestimation through pessimistic estimates \cite{zhang2024entropy} or improving optimization via actor-critic frameworks \cite{gao2025behavior}. 
    % Furthermore, recent efforts enhance generalization by integrating world models \cite{chandra2025diwa} or leveraging large language models (LLMs) for cross-task policy generation \cite{zhang2025llm}.
    
    \subsection{Generative Policy in Online RL}
    The development of generative policies in online RL has explored several model classes. Among these, diffusion models are increasingly prominent.
    Seminal work by Yang et al. \cite{yang2023policy} introduced the action gradient method, which combines action gradient updates with behavior cloning for online policy improvement.
    
    Research has rapidly evolved to address key challenges. To tackle training complexity, Psenka et al. \cite{DBLP:conf/icml/PsenkaEA024} developed Q-score matching (QSM), which aligns Q-function gradients with the score structure of the diffusion model to bypass full backpropagation. To enhance exploration and diversity, DACER \cite{wang2024diffusion} introduced an entropy regulation mechanism, while deep diffusion policy gradient (DDiffPG) \cite{li2024learning} utilized unsupervised clustering to mine diverse policy patterns. 
    Recent advancements have pushed performance and stability further. Reweighted Score Matching (RSM) \cite{DBLP:conf/icml/MaCW0025} yielded the diffusion policy mirror descent (DPMD) and soft diffusion actor-critic (SDAC) algorithms, achieving performance significantly surpassing traditional methods. Q-weighted variational policy optimization (QVPO) \cite{ding2024diffusion} similarly enhanced stability by combining Q-weighted variational lower bounds with entropy regularization.

    \section{Preliminary}
    \subsection{Online Reinforcement Learning}
    
    RL problems are typically modeled as a Markov decision process (MDP). Specifically, an MDP is defined by a quintuple
    \(\mathcal{M} = (\mathcal{S}, \mathcal{A}, P, R, \gamma)\)
    where \(\mathcal{S}\) denotes the state space, \(\mathcal{A}\) denotes the action space,
    \(P(s'|s,a)\) denotes the transition probability from state \(s\) to state \(s'\) after action \(a\),
    \(R(s,a)\) is the reward function,
    and \(\gamma \in (0,1)\) is the discount factor.
    The core objective of RL is to learn a policy \(\pi\)
    by maximizing the expected cumulative discounted reward,  
    \begin{equation}
    J(\pi) = 
    \mathbb{E}_{s_t \sim d^\pi,\, a_t \sim \pi(\cdot|s_t)} 
    \left[ \sum_{t=0}^{\infty} \gamma^t R(s_t, a_t) \right],
    \end{equation}
    where $d^{\pi}$ denotes the stationary state distribution induced by policy $\pi$. Maximizing this objective enables the agent to achieve long-term optimization within the environment.
    In pursuing this goal, the Q-function serves as a key tool for policy optimization,
    defined as
    \begin{equation}
    Q^\pi(s,a) = 
    \mathbb{E}_{s' \sim P(\cdot|s,a)} 
    \Big[ R(s,a) + \gamma \, 
    \mathbb{E}_{a' \sim \pi(\cdot|s')} [ Q^\pi(s',a') ] \Big].
    \end{equation}
    
    This equation is known as the Bellman expectation equation,
    describing the expected return for each state–action pair.
    In online RL, agents continuously learn and improve their policies through direct interaction with the environment. This process generates state transitions $(s_t, a_t, r_t, s_{t+1})$, which are typically stored in an experience replay buffer $\mathcal{B}$ to facilitate off-policy training.
    Through experience replay, the agent can revisit past interactions across different time steps, thereby enhancing the stability and performance of the policy.

    \subsection{Flow Matching}
    Flow matching \cite{DBLP:conf/iclr/LipmanCBNL23} is a generative modeling method proposed in recent years. Unlike diffusion models based on stochastic differential equations (SDEs) \cite{DBLP:conf/iclr/0011SKKEP21},
    it employs deterministic ordinary differential equations (ODEs) to characterize the gradual transformation from a noise distribution to a target data distribution,
    thereby simplifying the training objective and accelerating sampling. Its core lies in learning a time-dependent velocity field
    that drives samples along continuous trajectories from the prior noise distribution to the target distribution.
    
    Let $x_0 \sim p_{\text{prior}}(x)$ denote a noise sample from the prior, and $x_1 \sim p_{\text{data}}(x)$ denote a target data sample.
    Define linear interpolation over time $t \in [0,1]$:
    \begin{equation}
    x_t = (1-t)\,x_0 + t\,x_1 .
    \end{equation}
    Along this interpolation path, the corresponding target velocity field is
    \begin{equation}
        u(x_t) \triangleq \frac{dx_t}{dt} = x_1 - x_0 .
    \end{equation}
    
    The training objective of Flow Matching is to learn a neural network $v_\theta(x,t)$ to regress this target velocity field.
    The optimization problem is:
    \begin{equation}
    \mathcal{L}_{\text{FM}}(\theta)
    = \mathbb{E}_{\,x_0 \sim p_{\text{prior}},\, x_1 \sim p_{\text{data}},\, t \sim \mathcal{U}[0,1]}
    \bigl[ \,\|\, u(x_t) - v_\theta(x_t,t) \,\|_2^2 \,\bigr].
    \end{equation}
    
    Upon training convergence, \(v_\theta\) approximates the velocity field from the noise distribution to the data distribution.
    Thus, samples can be generated by numerically solving the following ODE:
    \begin{equation}
    \frac{d}{dt} z_t = v_\theta(z_t, t), \qquad z_0 \sim p_{\text{prior}}.
    \end{equation}
    
    Advancing the initial noise $z_0$ over time from $t=0$ to $t=1$ yields $z_1$ as a generated sample drawn from the target distribution $p_{\text{data}}$.
    In practice, sampling can be performed using the Euler method or a high-order ODE solver.
    
    % Compared to diffusion models, Flow Matching models the probability flow through deterministic vector fields,
    % reducing training and inference complexity while maintaining generation quality and significantly improving sampling efficiency.
    
    \section{Methodology}
    \subsection{Flow Models as Policy Representations}
    Based on the flow matching principle introduced in the previous section, we represent the policy $\pi_\theta(\cdot|s)$ as a conditional generative process. This process learns a velocity field $v_\theta(a, t, s)$ determined by the state $s$, which transforms a simple prior noise $a_0 \sim \mathcal{N}(0, I)$ into a high-value action $a_1$. This generative process (i.e., policy sampling) is defined by the integral of the following ODE: \begin{equation}a_1 = \pi_\theta(s) = a_0 + \int_{0}^{1} v_{\theta}(a_t, t, s) \text{dt}.\label{equ:flow_continuous_infer}\end{equation}
    
    In practice, this integral is solved numerically. For example, using $N$-step Euler method for discretization, the action sampling process is expressed as: \begin{equation}a_1 = a_0 + \frac{1}{N}\sum_{k=1}^{N}v_{\theta}(a_k,\frac{k}{N}, s).\label{equ:flow_distrete_infer}\end{equation} 
    
    To train this conditional velocity field $v_\theta$, we make it imitate a target vector field $u(a_t, s)$. As discussed in the previous section, under the linear interpolation path $a_t = (1-t)a_0 + t a_1$, this target vector field is $u(a_t, s) = a_1 - a_0$. Therefore, we train the policy network $\theta$ using the conditional flow matching (CFM) loss\cite{DBLP:conf/iclr/LipmanCBNL23}: \begin{equation}\mathcal{L}_\text{CFM}(\theta) = \mathbb{E}_{s, a_0, a_1, t} [ ||v_{\theta}(a_t, t, s) - (a_1 - a_0)||^2 ],\end{equation} where $a_1$ is an action sampled from some “target distribution” $p_\text{target}(\cdot|s)$. However, in online RL, this $p_\text{target}$ is unknown. How to construct this target distribution in online RL is precisely the core challenge addressed in the next section.

    \subsection{Action-Strengthening Flow Policy}
    As mentioned earlier, the greatest challenge in current flow strategies and online RL is the lack of a suitable target distribution. To address this issue, we propose a dynamic target guidance mechanism. Rather than seeking a fixed imitation target, we assume that the optimal policy distribution $\pi^*(a|s)$ can be approximated by an energy-based model implicitly defined by the Q-function:
    \begin{equation}
    p(a|s) \propto \exp(Q(s,a)/\alpha),
    \label{equ:alpha}
    \end{equation}
    where $\alpha$ is a temperature parameter. This distribution inherently assigns high probabilities to actions with high Q-values. 
    % Consequently, the training objective for the flow policy shifts to: how to imitate this energy distribution implicitly defined by the Q-function? A direct approach is to use the gradient of the Q-function, $\nabla_a Q(s,a)$, to find the peak of this distribution, as shown in (4).
    Consequently, the training objective for the flow policy is reformulated as the task of approximating the energy distribution implicitly defined by the Q-function. A straightforward approach leveraging the Q-function gradient, i.e., $\nabla_a Q(s,a)$, to seek the mode of this distribution, is as follows:
    \begin{equation}
    a\longleftarrow a+\nabla_aQ(s,a).
    \label{equ:dipo}
    \end{equation}
    
    However, in practice, we find that this pure gradient ascent approach \cite{yang2023policy} leads to overly deterministic policies that easily get stuck in narrow local distributions, neglecting the inherent randomness of the energy distribution itself. 
    To address this issue, we instead employ Langevin dynamics, a standard technique for sampling from energy models, to obtain samples $a^*$ from $p(a|s) \propto \exp(Q(s,a)/\alpha)$.
    This ensures our target $a^*$ both points toward high Q-values and preserves exploration. The specific sampling process is as follows:
    \begin{equation}
    a_{t}=a_{t-1}+\eta_a\nabla_{a} Q ( s, a_{t-1} )+\sqrt{2\eta_a\alpha}\xi ,
    \end{equation}
    where $\eta_a$ denotes the step size, $\xi$ is noise sampled from the Gaussian distribution $N(0,1)$, and $\alpha$ is the temperature parameter previously defined in Eq. (\ref{equ:alpha}).
    The specific algorithm implementation is shown in Algorithm \ref{alg:langevin}.
    
    % \begin{algorithm}[t]
    \begin{algorithm}
    \caption{Langevin Optimize Action}
    \label{alg:langevin}
    \begin{algorithmic}
    \Require state $s$, initial action $a_{\mathcal{B}}$ from buffer, step size $\eta_a$, temperature $\alpha$, number of steps $N_\ell$
    \State $a \gets a_{\mathcal{B}}$
    \For{$i = 1$ to $N_\ell$}
        \State Compute $Q\gets Q_{\phi}(s,a)$,
        \State Compute gradient $\nabla_a Q$
        \State Sample $\xi \sim \mathcal{N}(0,I)$
        \State $a \gets a + \eta_a \nabla_a Q + \sqrt{2\eta_a\alpha}\,\xi$
    \EndFor
    \Ensure optimized action $a^\star$
    \end{algorithmic}
    \end{algorithm}
    
    \subsection{Critic Learning with Double Q-networks}
    To mitigate overestimation bias, the critic component employs a dual Q-learning approach\cite{fujimoto2018addressing}. This method involves constructing and training two Q-networks, i.e., $Q_{\phi_1}(s,a)$ and $Q_{\phi_2}(s,a)$, to approximate Q-values. Furthermore, training stability is enhanced by employing target networks $Q_{\bar \phi_1}(s,a)$ and $Q_{\bar \phi_2}(s,a)$, which are soft copies of the Q-networks\cite{van2016deep}. The objective function for updating the Q-networks is derived from minimizing the Bellman error. For each Q-network $Q_{\phi_i}(s,a)$, the loss function $J_{Q} ( \phi_{i} )$ is given by the following formula:
    \begin{equation}
    \begin{aligned}
    J_{Q} ( \phi_{i} )
    &=\underset{\substack{
    (s, a, r, s^{\prime} ) \sim\mathcal{B} \\
    % a^{\prime} \sim\pi_{\mathrm{flow}}
    a^{\prime} \sim\pi_\theta(\cdot|s')
    }}{\mathbb{E}}
    \Big[
     (r + \gamma \operatorname* \min_{j=1, 2} Q_{\bar{\phi}_{j}} ( s^{\prime}, a^{\prime} ) - Q_{\phi_{i}} ( s, a ) )^{2}
    \Big],
    \label{eq:q_loss}
    \end{aligned}
    \end{equation}
    where the objective value is computed as the minimum between two target Q-networks, $Q_{\bar{\phi}_{1}} ( s^{\prime}, a^{\prime} )$ and $Q_{\bar{\phi}_{2}} ( s^{\prime}, a^{\prime} )$. % The distributional value function method introduced in DSAC\cite{duan2025distributional} is also employed to further mitigate overestimation.
    
    \subsection{Practical Algorithm Implementation}
    Our total actor loss function $J_F(\theta)$ is shown below, which is a hybrid objective.
    \begin{equation}
    \begin{aligned}
    J_F(\theta)
    &=\underset{\substack{
    a \sim \pi_{\mathrm{flow}} \\
    a_0 \sim \mathcal{N}(0,1)
    }}{\mathbb{E}}
    \Big[
    - Q(s, \pi_\theta(s)) \\
    &\quad + \lambda_f
    \| v_\theta(s, a_t, t) - (a^\star - a_0) \|_2^2
    \Big].
    \label{eq:flow_policy_loss}
    \end{aligned}
    \end{equation}
    
    This loss function combines policy improvement with guided imitation. This loss function consists of two core components: The first part is a policy gradient term ($-Q(s, \pi_\theta(s))$), a standard RL policy improvement objective used to directly increase the Q-value of actions output by the policy $\pi_\theta$ to maximize the expected return. The second part is a flow-matching imitation term ($\lambda_f \| v_\theta(s, a_t, t) - (a^\star - a_0) \|_2^2$), which trains the flow policy $v_\theta$ to imitate the optimized, `better' target action $a^\star$ generated by Algorithm \ref{alg:langevin}. Regarding the design of the dynamic weighting coefficient $\lambda_f$, to ensure training stability, we reference the approach from \cite{lv2025flow} and set $\lambda_f$ to an advantage-weighted form: $\lambda_f \propto \mathrm{ReLU}(Q(s,a^\star)-Q(s,a_{\mathcal{B}}))$, where $a_{\mathcal{B}}$ is the original action sampled from the replay buffer.
    The complete training procedure is shown in Algorithm \ref{alg:flowdipopi}.

    \begin{algorithm}[t]
    \caption{DACER-F}
    \label{alg:flowdipopi}
    % \begin{algorithmic}[1]
    \begin{algorithmic}
    \Require Discount factor $\gamma$, 
    learning rates $\beta_Q,\beta_F$, target smoothing coefficient $\tau$
    \State Initialize value networks $Q_{\phi_1}, Q_{\phi_2}$ and flow network $v_\theta$ 
    with random parameters $\phi_1, \phi_2, \theta$
    \State Initialize target networks: $\bar{\phi}_1 \gets \phi_1, \ \bar{\phi}_2 \gets \phi_2$
    \For{each iteration}
        \For{each interaction step}
            \State \textcolor{gray}{\textit{// Actor inference}}
            
            \State Sample $a \sim \pi_{\text{flow}}(\cdot|s)$ using (\ref{equ:flow_distrete_infer})
            \State Observe reward $r$ and new state $s'$
            \State Store the experience $(s,a,r,s')$ in the buffer $\mathcal{B}$
        \EndFor
        \State \textcolor{gray} {\textit{// Replay}}
        \State Sample a batch of data from $\mathcal{B}$
        \State \textcolor{gray} {\textit{// Critic update}}
        \State Update value networks by minimizing $J_Q(\phi_i)$
        \State \textcolor{gray} {\textit{// Actor update}}
        \State Update flow network by minimizing $J_F(\theta)$
        \State Update target networks: 
        $\bar{\phi}_i \gets (1-\tau)\bar{\phi}_i + \tau\phi_i$
    \EndFor
    \State {\textbf{return}} trained parameters $\theta, \phi_1, \phi_2$
    \end{algorithmic}
    \end{algorithm}

    \section{Experiment}
    % In this section, we conduct extensive simulations in two key types of procedurally generated driving scenarios \cite{jiang2023reinforcement}, namely multi-lane highways and urban crossroads, to demonstrate the effectiveness of the proposed DACER-F algorithm.
    In this section, we conduct extensive simulations in two key types of procedurally generated driving scenarios \cite{jiang2023reinforcement}, namely multi-lane highways and urban crossroads, to demonstrate the effectiveness of the proposed DACER-F algorithm. Subsequently, to verify its broad generalization capabilities, we evaluate the algorithm on the standard RL continuous control benchmark DMC.
    \subsection{Simulation Environment}
    
    We utilize two procedurally generated scenarios: a multilane highway for high-speed continuous driving and lane merging, and an urban intersection for complex turning interactions (straight, left, right, U-turn). Both scenarios support three switchable traffic densities (``sparse", ``normal", ``dense") with passenger cars as participants. 
    % The action space $\mathcal{A}$ is composed of acceleration and angular velocity increments, represented as $a = [\Delta a_x, \Delta \delta]$.
    The action space is composed of acceleration and angular velocity increments, represented as $a = [\Delta a_x, \Delta \delta]$.
    The observation space $S$ concatenates three components: (1) the ego-vehicle state $o_{\text{ego}} = [v_x, v_y, r, a_x, \delta]$; (2) the reference trajectory $o_{\text{ref}} = [x, y, \cos(\varphi), \sin(\varphi)] \times W$, where $W$ is the number of waypoints; and (3) surrounding vehicles $o_{\text{sur}} = [x_{\text{sur}}, y_{\text{sur}}, \cos(\varphi_{\text{sur}}), \sin(\varphi_{\text{sur}}), v_{\text{sur}}, \text{mask}] \times M$, where $M$ is the vehicle limit and $\text{mask}$ (1/0) indicates validity.
    
    The reward function balances trajectory tracking, control cost, ride comfort, and driving safety. The tracking reward is $R_{\text{Tracking}} = -\rho_{\text{lat}} \cdot h_{\text{lat}}(e_{\text{lat}}) - \rho_{\text{long}} \cdot h_{\text{long}}(e_{v}) - \rho_{\phi} \cdot h_{\phi}(e_{\phi})$, where $h(\cdot)$ is a Huber-like penalty. The control cost $R_{\text{Control}} = -\rho_{\text{acc}} \cdot |a_x|^3 - \rho_{\text{steer}} \cdot |\delta - \delta_{\text{nom}}|$ uses a cubic penalty to suppress extreme inputs. The comfort reward $R_{\text{Comfort}} = -\rho_{\text{yaw}} \cdot h_{\text{yaw}}(r) - \rho_{\text{jerk}} \cdot h_{\text{jerk}}(\dot{a}_x) - \rho_{\Delta \text{steer}} \cdot h_{\Delta \text{steer}}(\dot{\delta})$ penalizes abrupt dynamics. A liveness reward $r_{\text{live}} = \rho_{\text{step}} \cdot \frac{1}{3} \cdot \text{clip}(v_x, 0, 3.0)$ encourages progress.
    
    Additionally, safety-related costs include: safe following distance $c_{\text{front}} = \rho_{\text{front}} \cdot (1 - \tanh(x'_{\text{sur}} / (v_x \cdot \Delta t_{\text{safe}})))$; safe space costs ($c_{\text{space}}$, $c_{\text{side}}$, $c_{\text{rear}})$ for blind spots and improper maneuvers; and a boundary cost $c_{\text{boundary}} = \rho_{\text{boundary}} \cdot \max(0, d_{\text{bound}} - d_{\text{thresh}} + 1.0)$. Severe events like collisions and going off-road incur large penalties (e.g., 200, or dynamically calculated). The mentioned reward and cost weights are listed in Table \ref{tab:hyperparameters}. All values are in international units by default.

    % \begin{table}[t]
    % \caption{Hyper-parameters of reward and cost.}
    % \centering
    % \renewcommand{\arraystretch}{1.2}
    % \begin{tabular}{cl}
    % \hline\hline
    % \textbf{Symbol} & \textbf{Value} \\
    % \hline
    % $\Delta a_x$ & [-0.4, 0.25] \\
    % $\Delta \delta$ & [-0.025, 0.025] \\
    % $a_x$ & [-0.8, 0.8] \\
    % $\delta$ & [-0.571, 0.571] \\
    % \hline
    % $\rho_\text{lat}$ & 1.5 \\
    % $\rho_\text{long}$ & 3.0 \\
    % $\rho_{\phi}$ & 0.3 \\
    % $\rho_\text{acc}$ & 1.0 \\
    % $\rho_\text{steer}$ & 0.1 \\
    % $\rho_\text{yaw}$ & 0.7 \\
    % $\rho_\text{jerk}$ & 0.5 \\
    % $\rho_{\Delta \text{steer}}$ & 0.005 \\
    % $\rho_\text{step}$ & 1.0 \\
    % \hline
    % $M$ & 8 \\
    % $\Delta t_\text{safe}$ & 1.0 \\
    % $\rho_\text{front}$ & 10.0 \\
    % $\rho_\text{side}$ & 0.0 \\
    % $\rho_\text{space}$ & 2.0 \\
    % $\rho_\text{rear}$ & 0.0 \\
    % $\rho_\text{boundary}$ & 20.0 \\
    % % $d_\text{bound}$ & 50.0 \\
    % $d_\text{thresh}$ & 0.5 \\
    % $N_l$ & 20 \\
    % % $d_{st}$ & 12.0 \\
    % % $d_{ss}$ & 2.0 \\
    % % $d_b$ & 1.8 \\
    % % $\lambda$ & 1.0 \\
    % \hline\hline
    % \end{tabular}
    % \label{tab:hyperparameters}
    % \end{table}
    
    \begin{table}[h]
    \caption{Hyper-parameters of reward and cost.}
    \centering
    \renewcommand{\arraystretch}{1.2}
    \begin{tabular}{cl | cl}
    \hline\hline
    \textbf{Symbol} & \textbf{Value} & \textbf{Symbol} & \textbf{Value} \\
    \hline
    $\Delta a_x$ & [-0.4, 0.25] & $\Delta \delta$ & [-0.025, 0.025] \\
    $a_x$ & [-0.8, 0.8] & $\delta$ & [-0.571, 0.571] \\
    \hline
    $\rho_\text{lat}$ & 1.5 & $\rho_\text{long}$ & 3.0 \\
    $\rho_{\phi}$ & 0.3 & $\rho_\text{acc}$ & 1.0 \\
    $\rho_\text{steer}$ & 0.1 & $\rho_\text{yaw}$ & 0.7 \\
    $\rho_\text{jerk}$ & 0.5 & $\rho_{\Delta \text{steer}}$ & 0.005 \\
    $\rho_\text{step}$ & 1.0 & & \\
    \hline
    $M$ & 8 & $\Delta t_\text{safe}$ & 1.0 \\
    $\rho_\text{front}$ & 10.0 & $\rho_\text{side}$ & 0.0 \\
    $\rho_\text{space}$ & 2.0 & $\rho_\text{rear}$ & 0.0 \\
    $\rho_\text{boundary}$ & 20.0 & $d_\text{thresh}$ & 0.5 \\
    $N_l$ & 20 & & \\
    \hline\hline
    \end{tabular}
    \label{tab:hyperparameters}
    \end{table}
    
    \subsection{Training Settings}
    % 我们使用GOPS (General Optimal control Problem Solver) 软件进行强化学习训练，主要的超参数列在表\ref{tab:train_hyperparameters}中。
    
    % 采样过程确保了场景的均衡分布。自车的初始状态是从交通流中随机选择的，并可施加一个小的随机状态偏移。参考轨迹是从当前或相邻车道中随机选择的，并带有用于跟踪的随机期望速度。周围车辆同样可以具有随机的位置偏移。在经验回放阶段，我们使用标准的经验回放方法 (Standard Experience Replay)。训练过程中产生的所有经验（状态、动作、奖励等）都会被存储在一个固定大小的回放缓存区中。在每次训练迭代时，会从该缓存区中进行均匀随机采样 (uniform random sampling)，抽取一个批次 (batch) 的数据用于更新策略网络。
    
    We conduct all RL training using the GOPS software\cite{wang2023gops}; key hyperparameters are detailed in Table \ref{tab:train_hyperparameters}. To ensure a balanced distribution of scenarios, we apply randomization during state initialization. The initial state of the ego-vehicle is randomly selected from the traffic flow with a potential small offset. Similarly, the reference trajectory is chosen from current or adjacent lanes with a random target speed, and surrounding vehicles are assigned random positional offsets. 
    For network optimization, we employ the relativistic adaptive gradient descent (RAD)\cite{lyu2025conformal} optimizer.
    We employ a standard experience replay method: all experiences $(s, a, r, s')$ are stored in a fixed-size replay buffer, and a batch of data is uniformly sampled from this buffer at each training iteration to update the networks.
    
    \begin{table}[t]
    \caption{Training hyper-parameters.}
    \centering
    \renewcommand{\arraystretch}{1.2}
    \begin{tabular}{ll}
    \hline\hline
    \textbf{Symbol} & \textbf{Value} \\
    \hline
    Actor learning rate $\beta_F$ & $1 \times 10^{-4}$ \\
    Two Critic learning rate $\beta_Q$ & $1 \times 10^{-4}$ \\
    Sample batch size & 200 \\
    Replay batch size & 512 \\
    Optimizer & RAD \\
    Discount factor & 0.99 \\
    Number of iterations & $5 \times 10^{5}$ \\
    Policy update frequency & 3 \\
    Target update rate $\tau$ & 0.001 \\
    Langevin step size $\eta_a$ & 0.03 \\
    Langevin temperature $\alpha$ & 0.01 \\
    Policy sampling steps & 1 \\
    \hline\hline
    \end{tabular}
    \label{tab:train_hyperparameters}
    \end{table}

    % \subsection{Training Curves and Visualization Demos}
    \subsection{Policy Performance}
    To validate DACER-F, we compare it against two SOTA baselines: DSAC (a distributional algorithm with a unimodal Gaussian policy) and DACER (a diffusion-based online policy, configured with 20 sampling steps for optimal performance). We evaluate total average reward (TAR), arrival rate, and collision rate, reporting the mean and 95\% confidence intervals across 5 random seeds.

\begin{figure}[h] 
    \centering
    \includegraphics[width=0.9\linewidth]{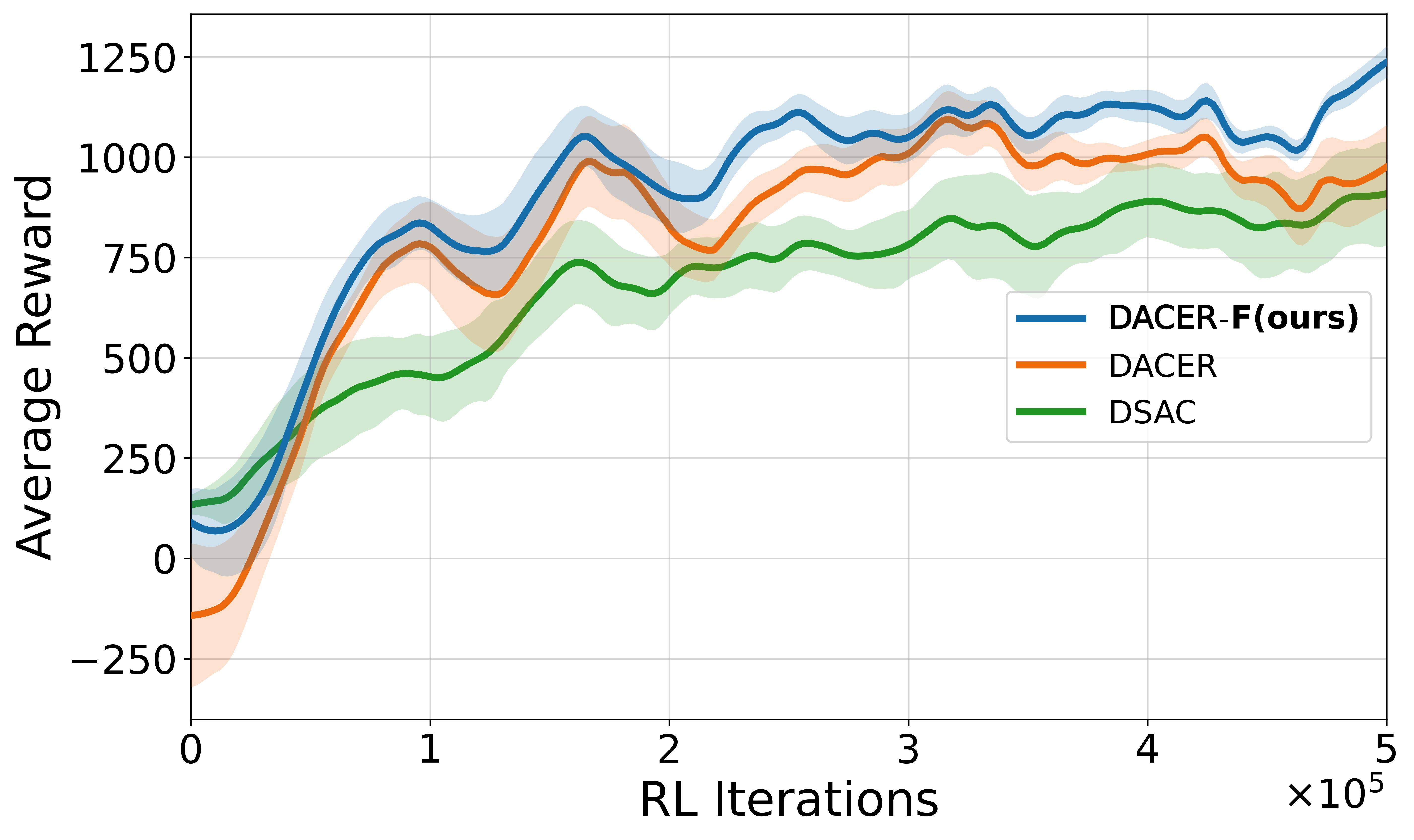}  % 图片路径和宽度
    \caption{Total average reward comparison during training.}  % 图片标题
    \label{fig:TAR}  % 图片引用标签
    \end{figure}

    As shown in Fig.~\ref{fig:TAR}, DACER-F exhibits rapid convergence and superior sample efficiency, consistently outperforming both baselines. Upon convergence, DACER-F achieves a final TAR of 1238, which is 28.0\% and 34.0\% higher than DACER (967) and DSAC (924), respectively. This highlights our method's exceptional capability in learning high-return policies.
    
    \begin{figure}[h] 
    \centering
    \includegraphics[width=0.9\linewidth]{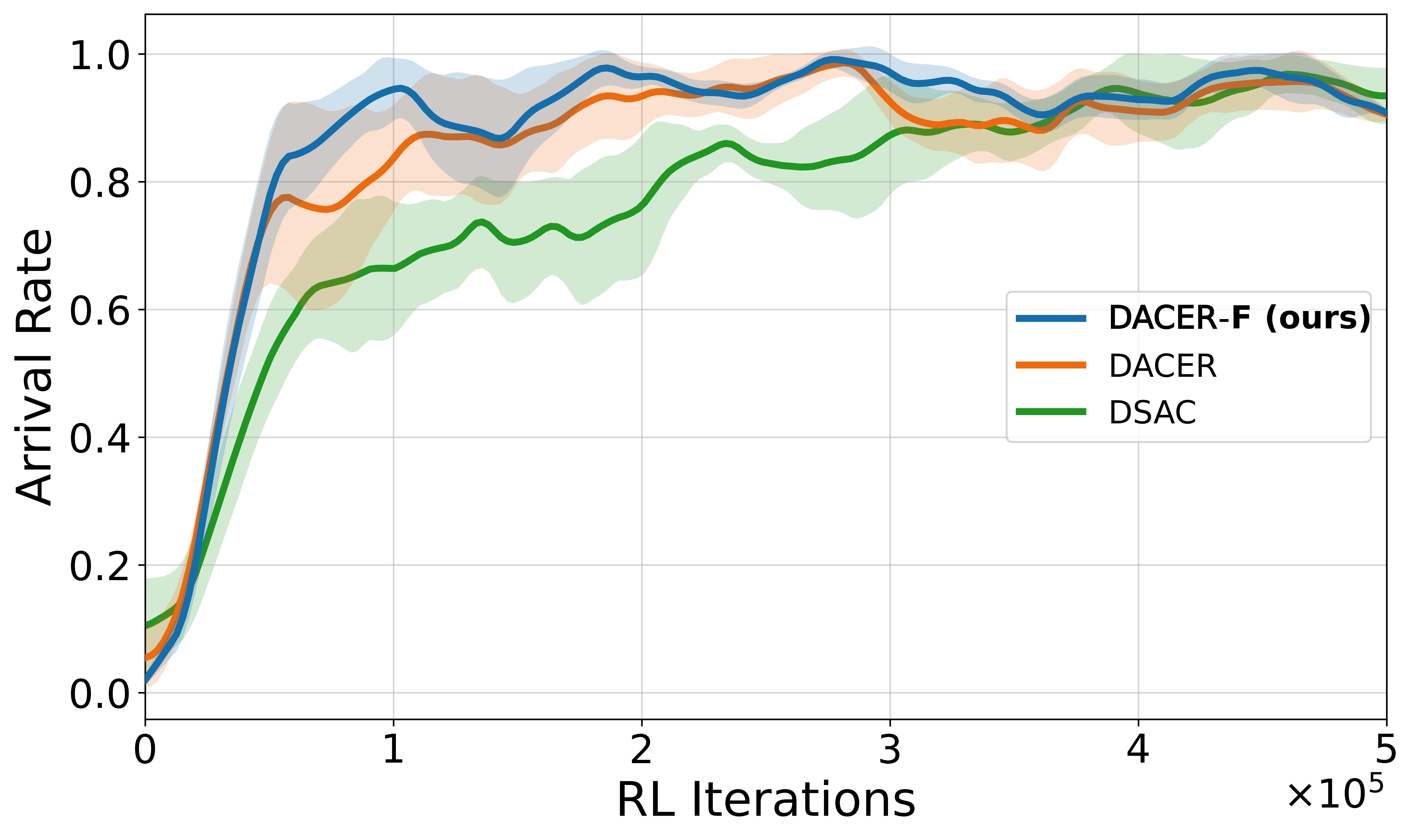}  % 图片路径和宽度
    \caption{Arrival rate comparison during training.}  % 图片标题
    \label{fig:arrival}  % 图片引用标签
    \end{figure}
    
    % Fig.~\ref{fig:arrival} highlights the significant advantage of our method in learning efficiency. Its arrival rate climbs the fastest, reaching and stabilizing at a near-optimal performance level after only approximately $1\times 10^5$ iterations. DACER also demonstrates strong, rapid convergence, performing very similarly to our method but stabilizing slightly slower. In contrast, while DSAC eventually converges to a similar high-performance level, its learning process is considerably slower than both generative policies (our method and DACER). This demonstrates the clear superiority of our method in convergence speed compared to DSAC and a slight efficiency edge over DACER.
    
    % Furthermore, Fig.~\ref{fig:collision} illustrates the safety performance. After a brief initial exploration phase (around 25,000 iterations), both our method and DACER converge to a similarly low collision rate, performing significantly better than DSAC for the vast majority of the training. Notably, our method demonstrates better initial stability, as its starting collision peak is substantially lower than the large initial spike seen in DACER. This reflects the ability of our method to quickly learn a safe policy while maintaining high data efficiency and stability.
    As shown in Fig.~\ref{fig:arrival}, our method exhibits superior learning efficiency, with its arrival rate stabilizing near-optimal after just $1\times 10^5$ iterations. This convergence is slightly faster than DACER and significantly outpaces DSAC. 
    
\begin{figure}[h] 
    \centering
    \includegraphics[width=0.9\linewidth]{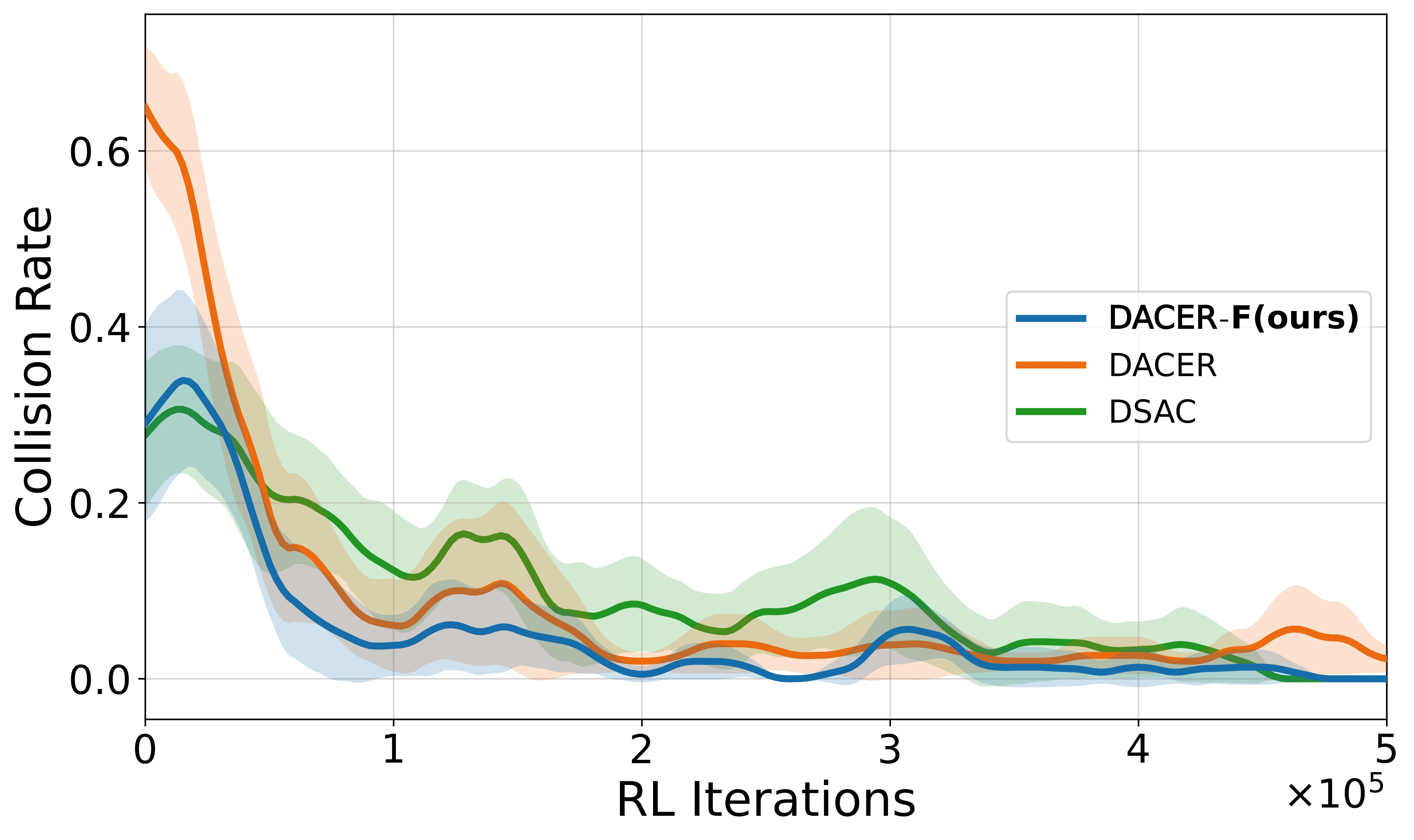}  % 图片路径和宽度
    \caption{Collision rate comparison during training.}  % 图片标题
    \label{fig:collision}  % 图片引用标签
    \end{figure}

    Furthermore, Fig.~\ref{fig:collision} illustrates that both our DACER-F and DACER achieve low collision rates after a brief 25,000-iteration exploration phase, consistently outperforming DSAC. Crucially, our method avoids the large initial collision spike observed in DACER, demonstrating exceptional early-stage stability and rapid safe policy learning.
    
    In summary, the experimental results clearly demonstrate that compared to DSAC and DACER, DACER-F achieves a superior trade-off between task performance and safety. It successfully learns to significantly improve task efficiency while simultaneously and substantially reducing safety risks.

    \subsection{Visualization Demonstrations}
\begin{figure*}[t]
        \centering % 将图表居中
        \captionsetup[subfigure]{font=footnotesize} 
        % (a)
        \subfloat[\small$t = 3.0\;\text{s}$]{\includegraphics[width=0.19\linewidth]{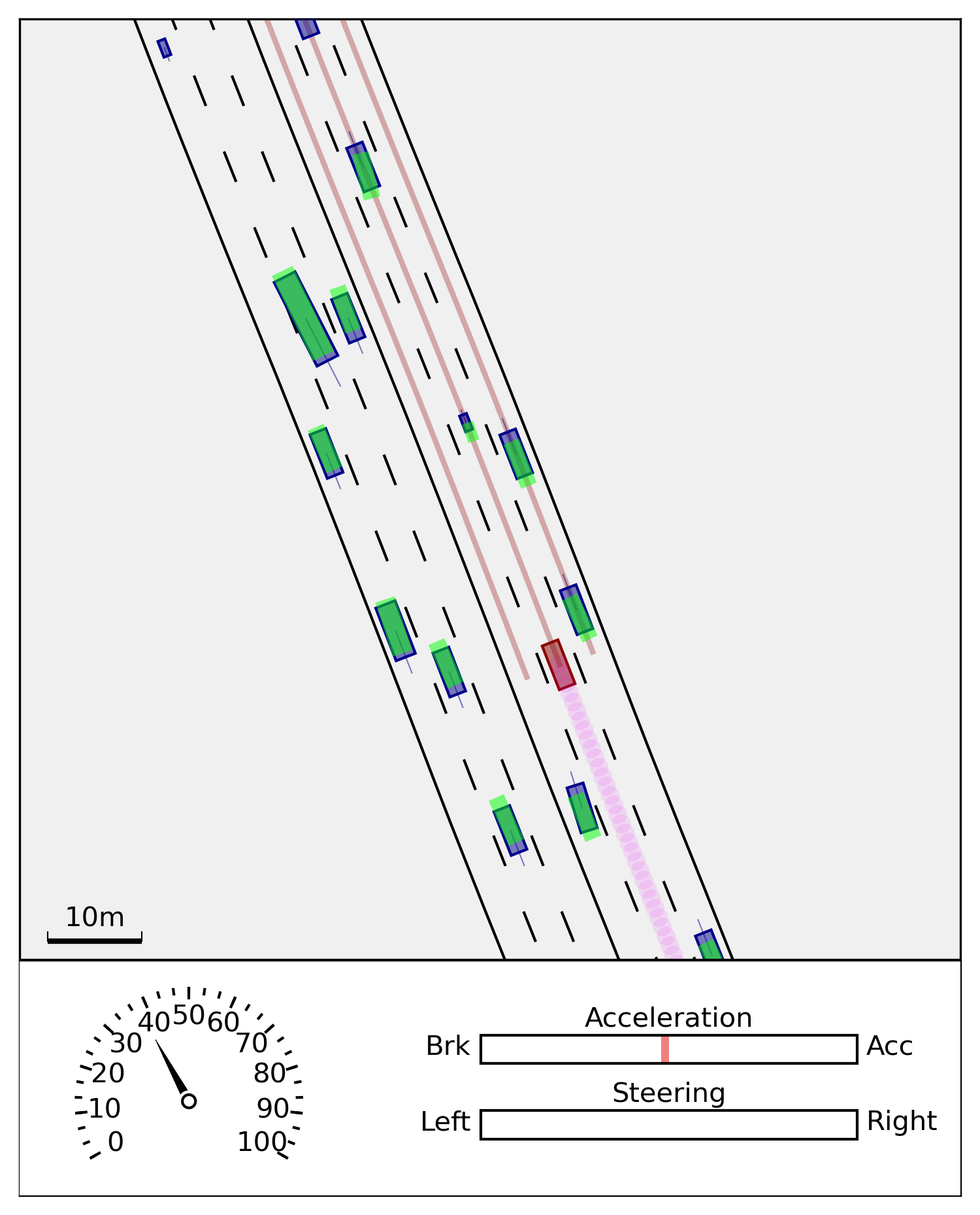}
        \label{fig:multilane_sub_a}}
        % \hfill % 添加水平间距
        % --- 警告：下面这一行必须是注释或代码，不能是空行 ---
        % (b)
        \subfloat[\small$t = 4.0\;\text{s}$]{\includegraphics[width=0.19\linewidth]{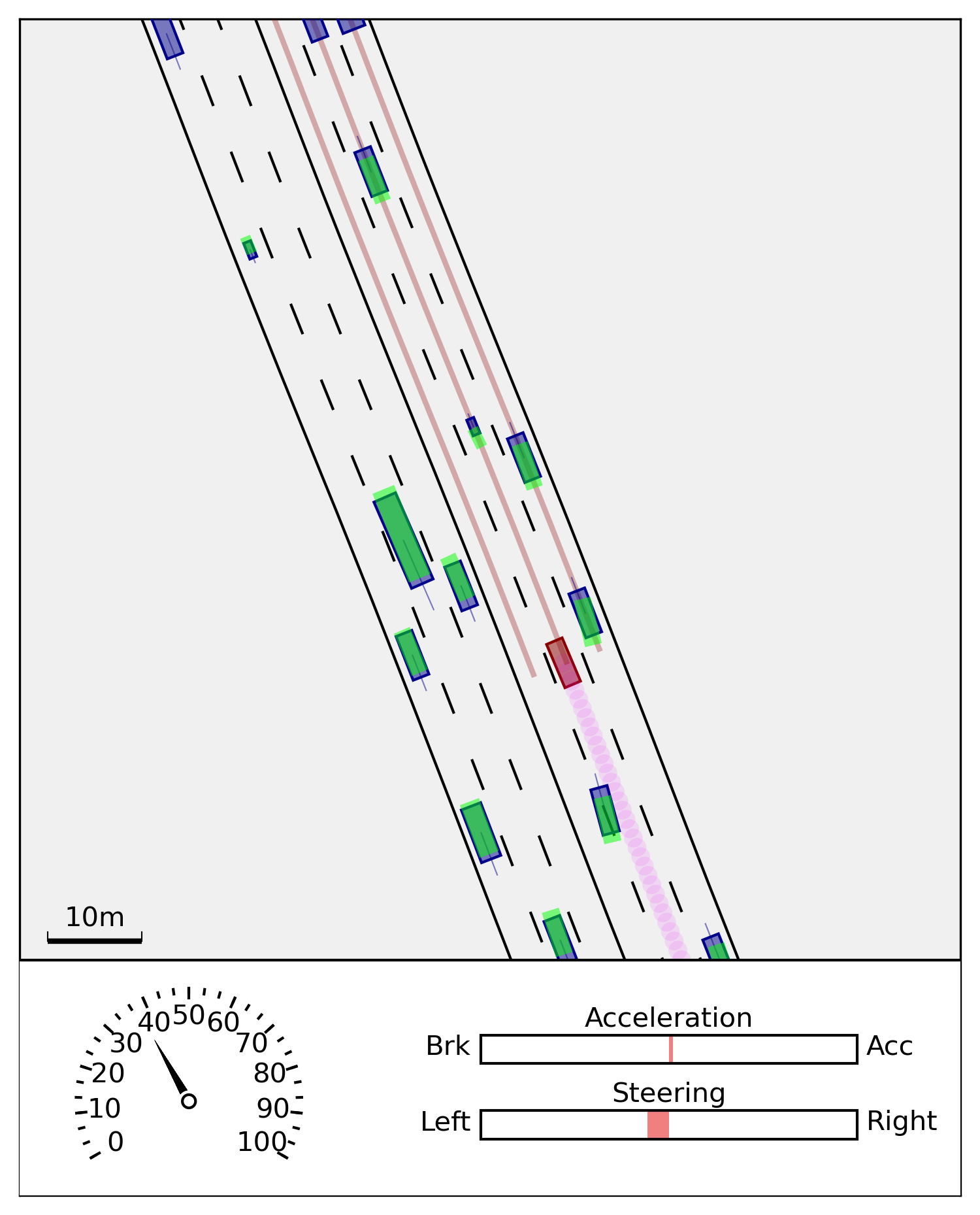}
        \label{fig:multilane_sub_b}} % <--- 已修正括号和 \label 位置
        % \hfill % 添加水平间距
        % --- 警告：下面这一行必须是注释或代码，不能是空行 ---
        % (c)
        \subfloat[\small$t = 6.5\;\text{s}$]{\includegraphics[width=0.19\linewidth]{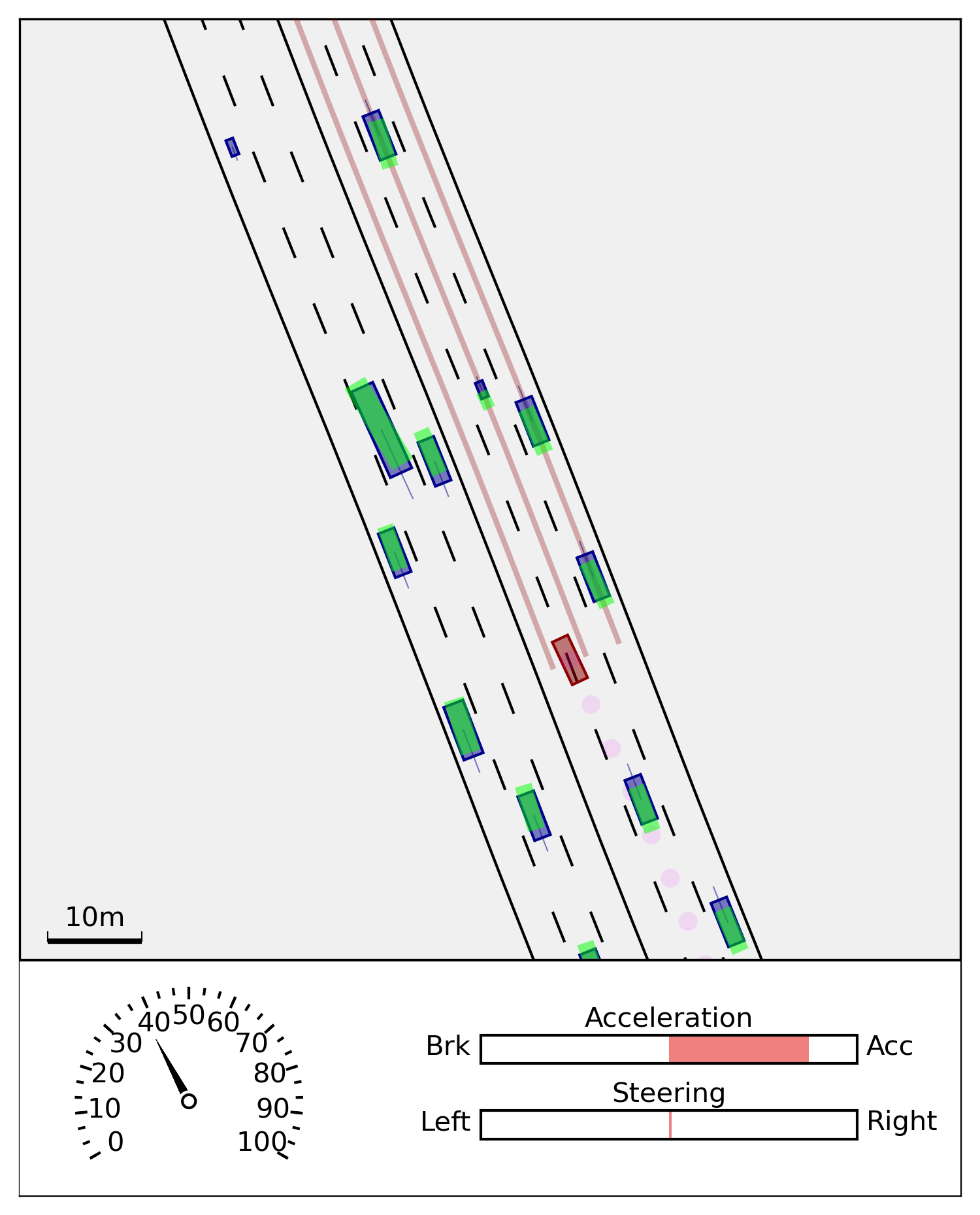}
        \label{fig:multilane_sub_c}} % <--- 已修正括号和 \label 位置
        % \hfill % 添加水平间距
        % --- 警告：下面这一行必须是注释或代码，不能是空行 ---
        % (d)
        \subfloat[\small$t = 9.5\;\text{s}$]{\includegraphics[width=0.19\linewidth]{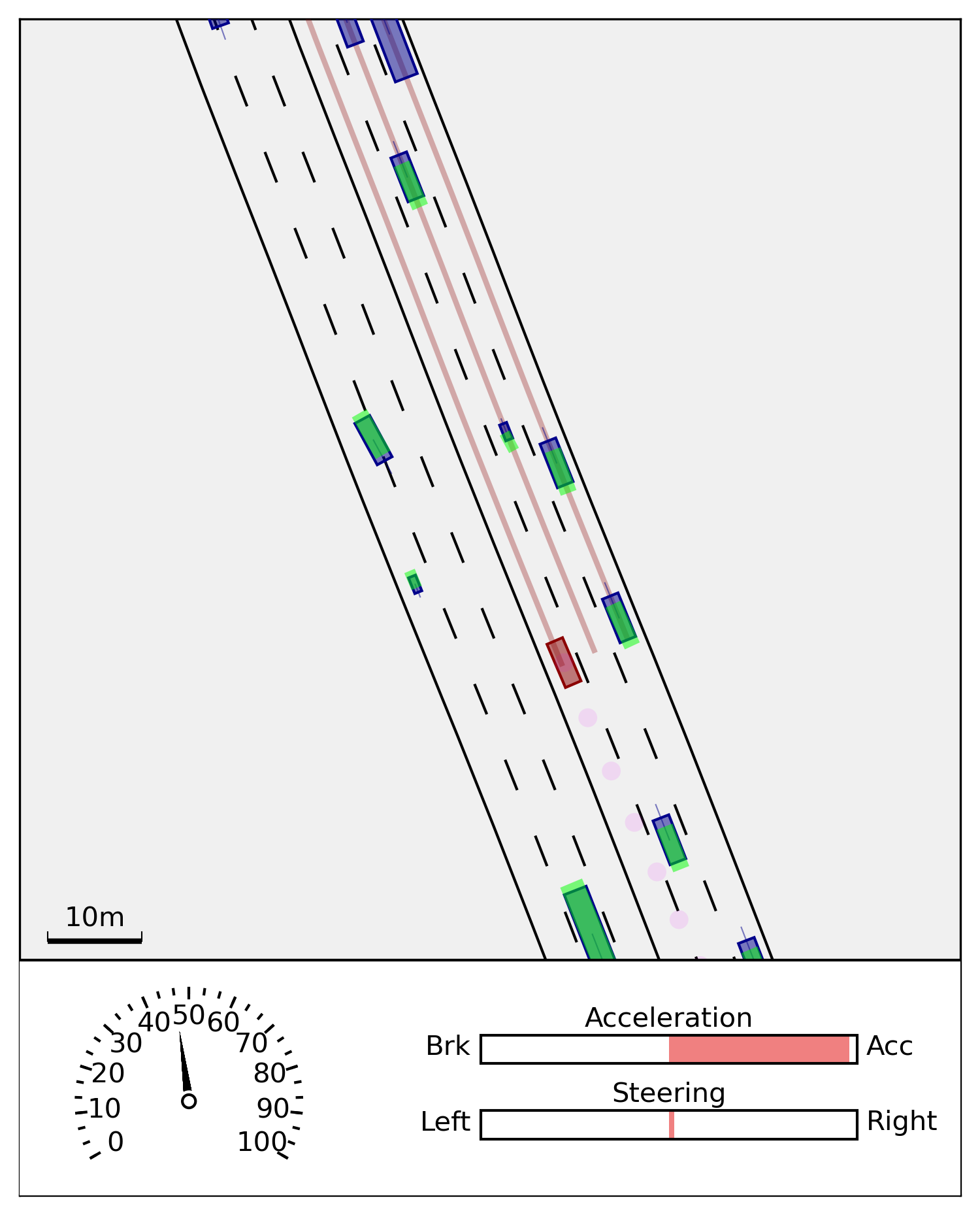}
        \label{fig:multilane_sub_d}} % <--- 已修正括号和 \label 位置
        % \hfill % 添加水平间距
        % --- 警告：下面这一行必须是注释或代码，不能是空行 ---
        % (e)
        \subfloat[\small$t = 12.5\;\text{s}$]{\includegraphics[width=0.19\linewidth]{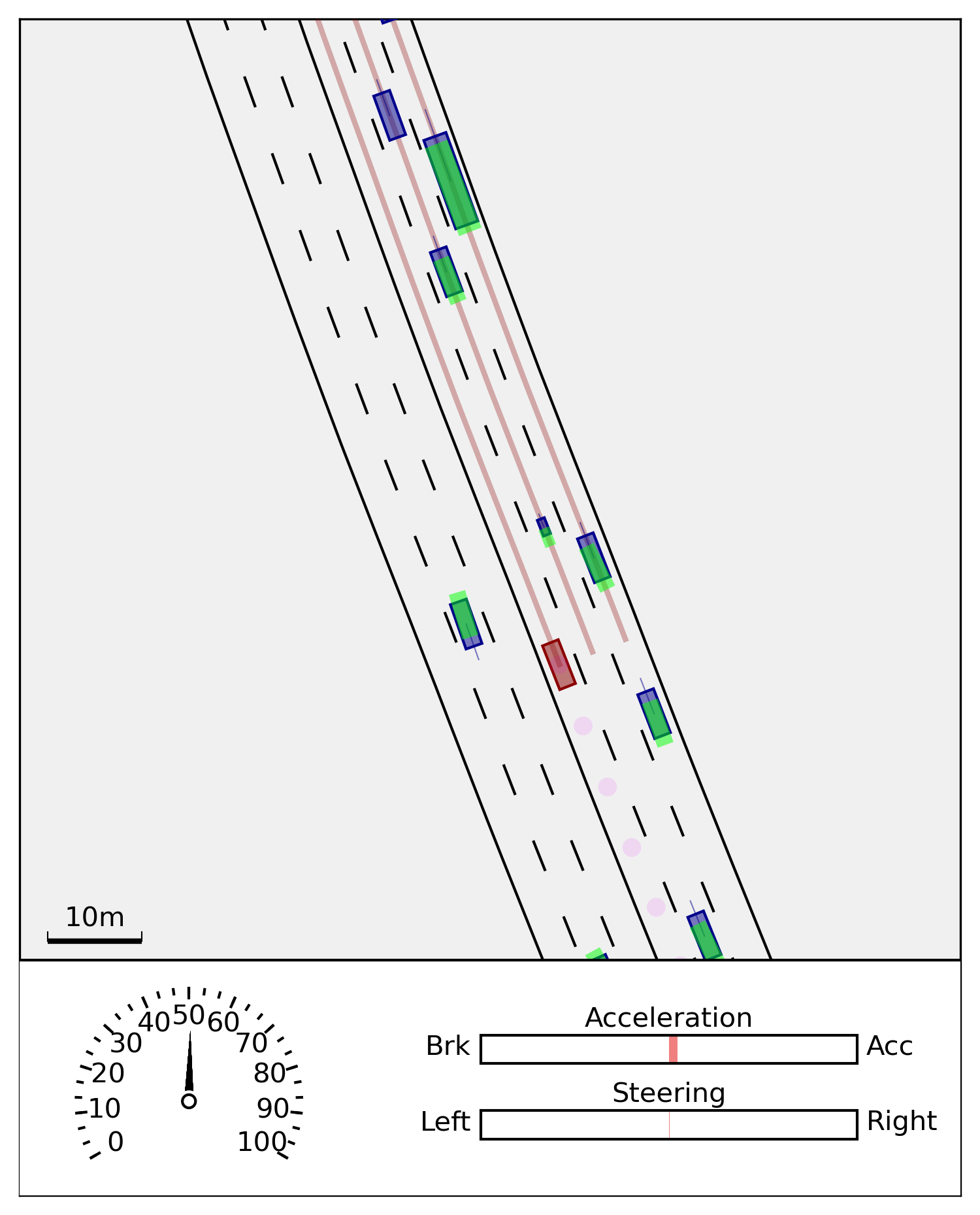}
        \label{fig:multilane_sub_e}} % <--- 已修正括号和 \label 位置
        % 整个图表的总标题
        \caption{Visualization of an autonomous lane change maneuver executed by the DACER-F agent (red vehicle) in the multilane scenario. The sequence (a-e) shows the agent smoothly merging into the left lane to overtake a slower vehicle.}
        \label{fig:multilane_main}
    \end{figure*}

\begin{figure*}[t]
        \centering % 将图表居中
        \captionsetup[subfloat]{font=footnotesize}
        % (a)
        \subfloat[\small$t = 17.5\;\text{s}$]{\includegraphics[width=0.19\linewidth]{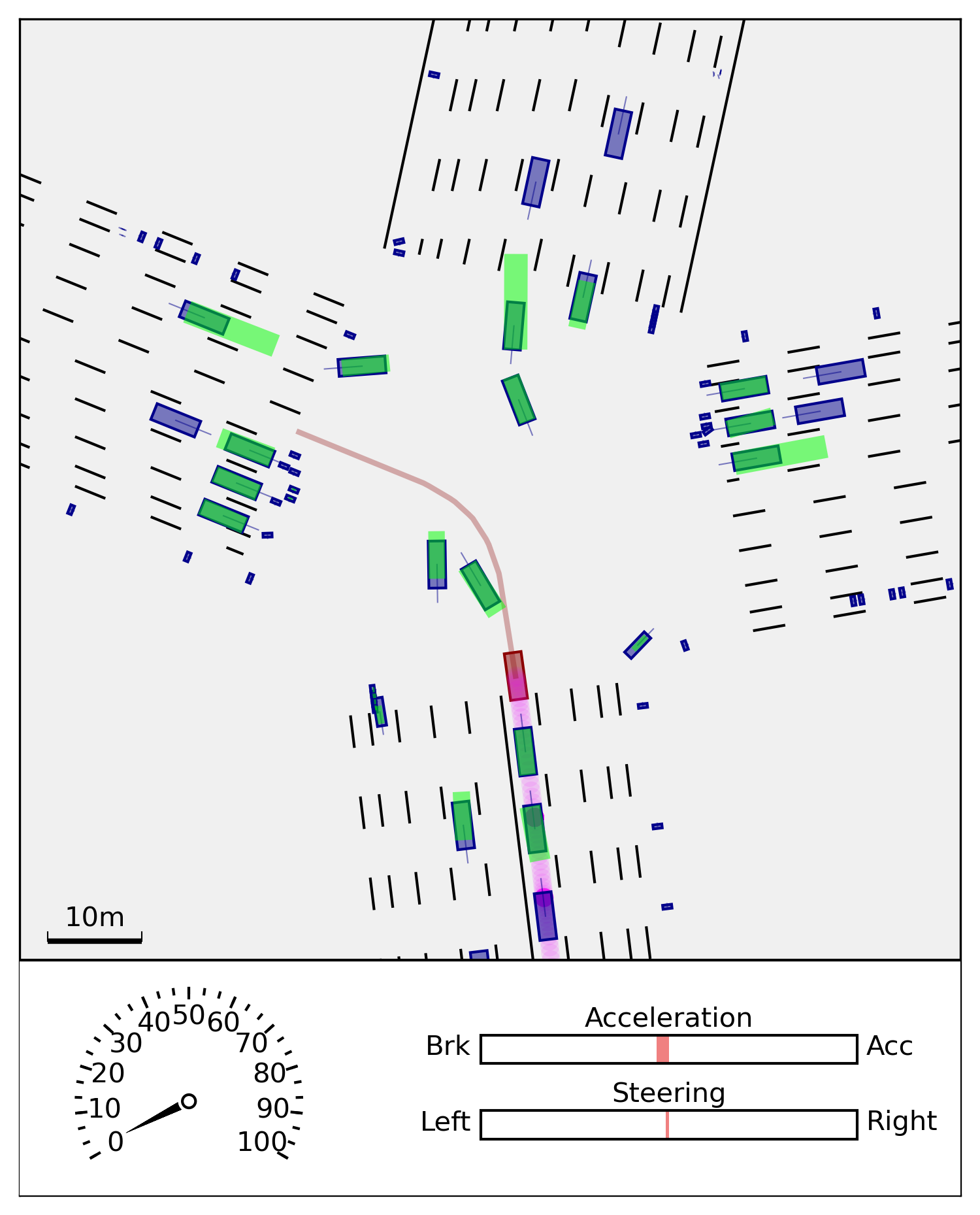}
        \label{fig:cross_sub_a}}
        % \hfill % 添加水平间距
        % --- 警告：下面这一行必须是注释或代码，不能是空行 ---
        % (b)
        \subfloat[\small$t = 21.0\;\text{s}$]{\includegraphics[width=0.19\linewidth]{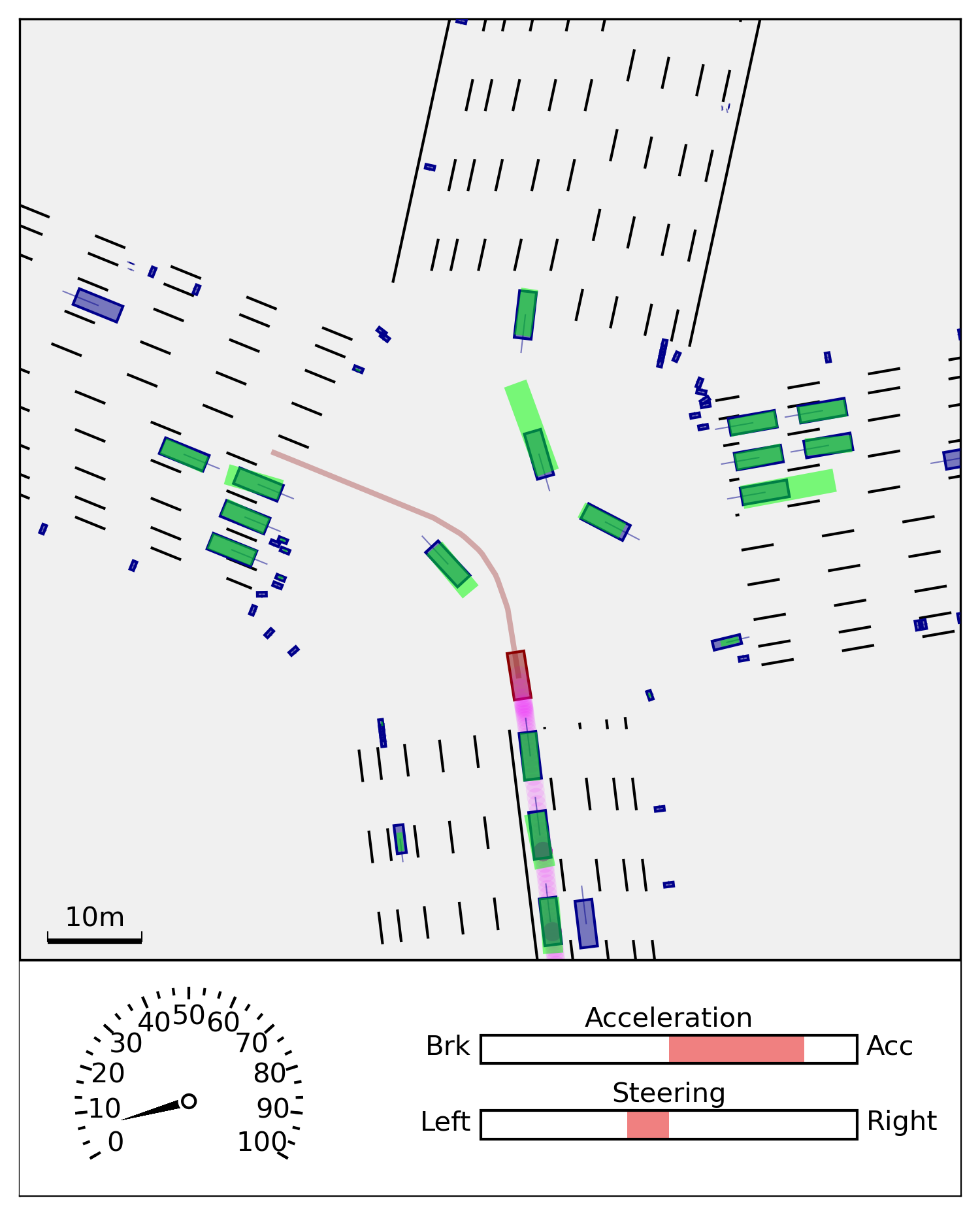}
        \label{fig:cross_sub_b}} % <--- 已修正括号和 \label 位置
        % \hfill % 添加水平间距
        % --- 警告：下面这一行必须是注释或代码，不能是空行 ---
        % (c)
        \subfloat[\small$t = 25.0\;\text{s}$]{\includegraphics[width=0.19\linewidth]{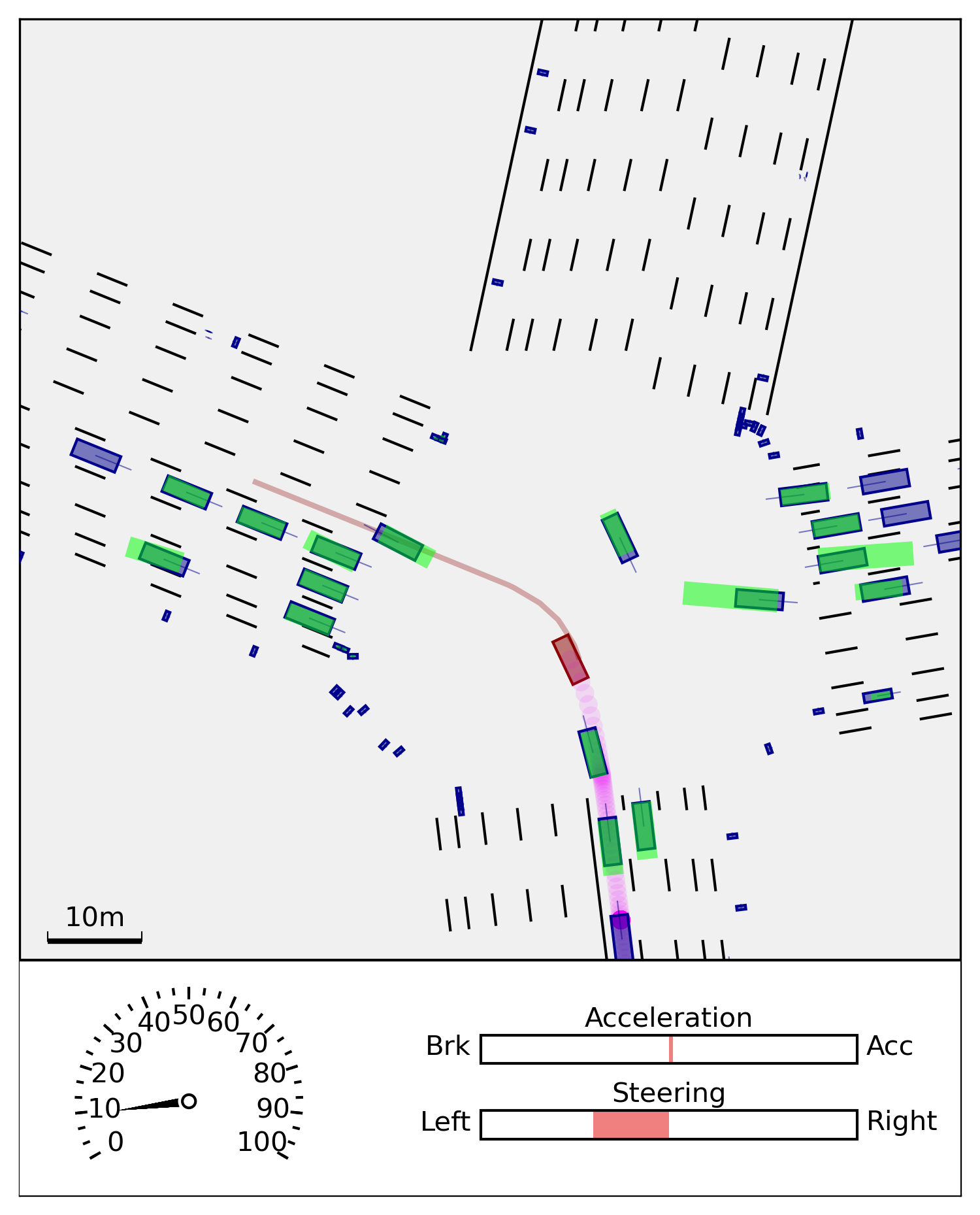}
        \label{fig:cross_sub_c}} % <--- 已修正括号和 \label 位置
        % \hfill % 添加水平间距
        % --- 警告：下面这一行必须是注释或代码，不能是空行 ---
        % (d)
        \subfloat[\small$t = 28.5\;\text{s}$]{\includegraphics[width=0.19\linewidth]{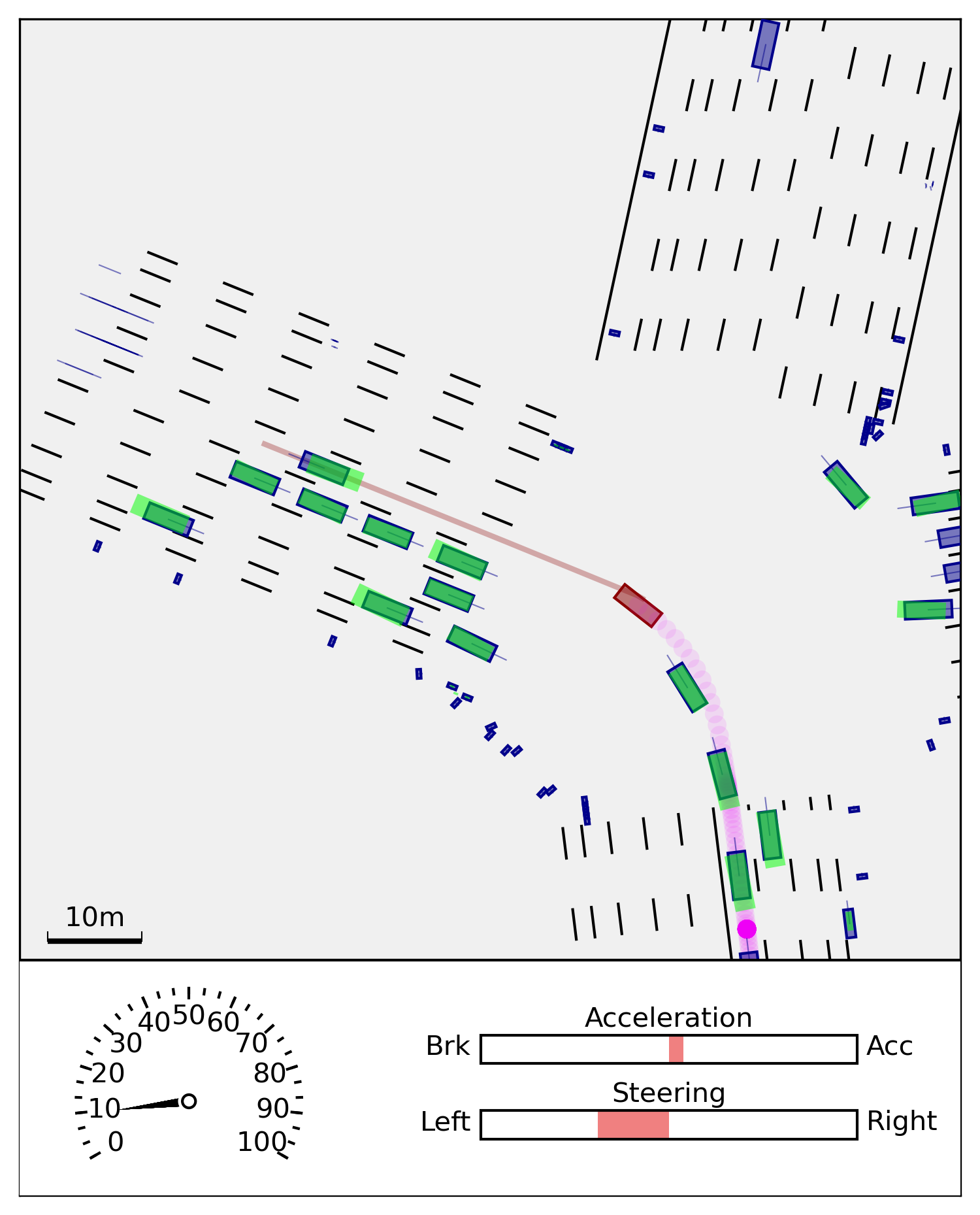}
        \label{fig:cross_sub_d}} % <--- 已修正括号和 \label 位置
        % \hfill % 添加水平间距
        % --- 警告：下面这一行必须是注释或代码，不能是空行 ---
        % (e)
        \subfloat[\small$t = 34.5\;\text{s}$]{\includegraphics[width=0.19\linewidth]{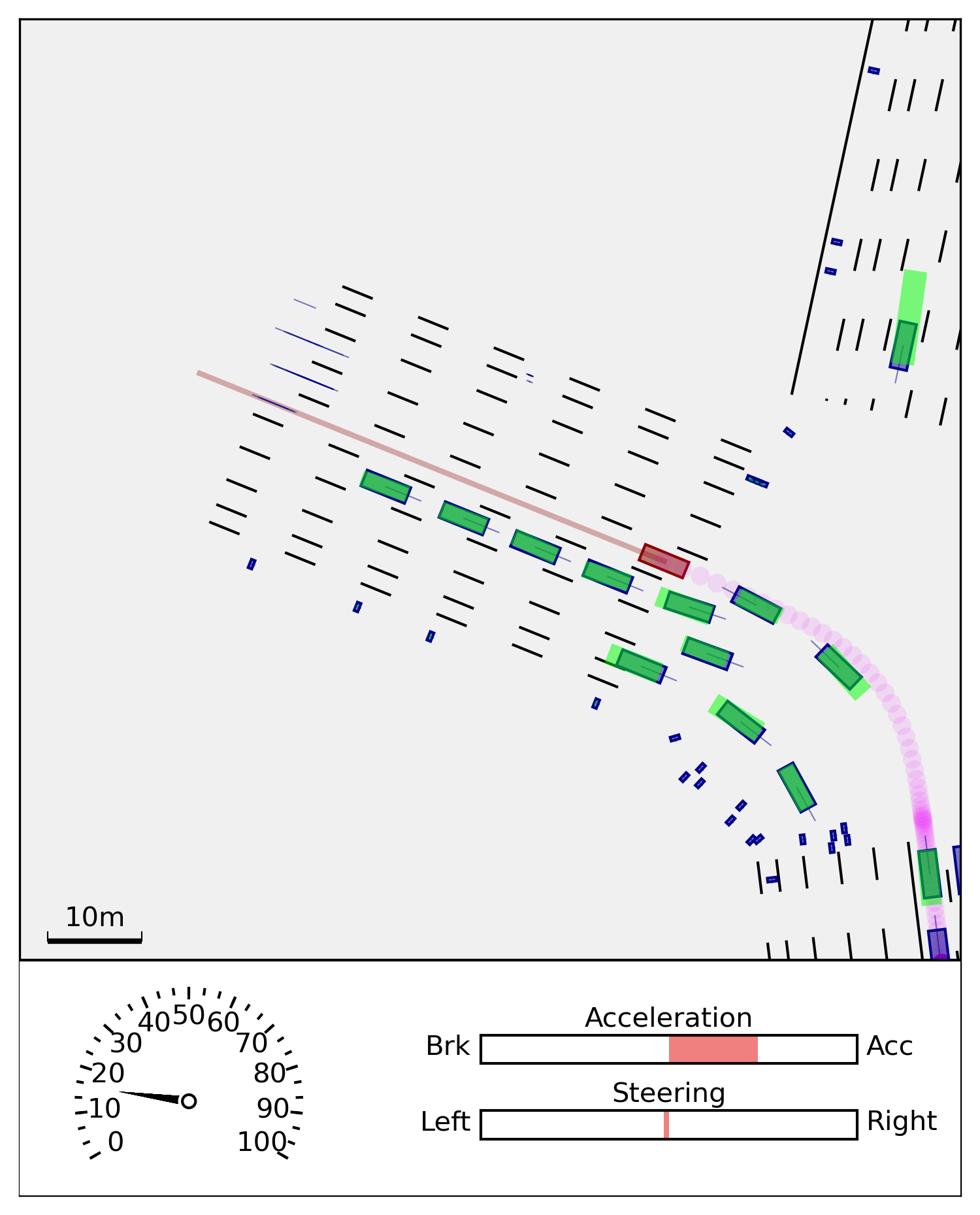}
        \label{fig:cross_sub_e}} % <--- 已修正括号和 \label 位置
        % 整个图表的总标题
        \caption{Visualization of a complex left-turn maneuver at the crossroad. The DACER-F agent (red vehicle) demonstrates social awareness by waiting for oncoming traffic (a), identifying a safe gap (b), and then proceeding to cross the intersection (c-e).}
        \label{fig:crossroad_main}
    \end{figure*}

    To qualitatively evaluate the performance of the DACER-F policy, Fig.~\ref{fig:multilane_main} and Fig.~\ref{fig:crossroad_main} present simulation cases in two typical driving scenarios. Fig.~\ref{fig:multilane_main} visualizes an autonomous lane change in the multi-lane scenario, demonstrating a complete and smooth maneuver. At $t=3.0\;\text{s}$ (Fig.~\ref{fig:multilane_main}a), the agent (red vehicle) observes the vehicle ahead and decides to execute a lane change. Subsequently, at $t=4.0\;\text{s}$ (Fig.~\ref{fig:multilane_main}b), the policy begins to steer smoothly and initiate the maneuver. During the $t=6.5\;\text{s}$ to $t=9.5\;\text{s}$ period (Figs.~\ref{fig:multilane_main}c, d), the agent accelerates steadily and safely merges into the left lane. Finally, at $t=12.5\;\text{s}$ (Fig.~\ref{fig:multilane_main}e), the vehicle has resumed stable cruising in the new lane after reaching the set reference speed.

    % 图\ref{fig:crossroad_main}可视化了DACER-F算法在十字路口与对向车流的复杂交互。在$t=17.5s$时(图\ref{fig:cross_sub_a})，智能体（红色车辆）抵达路口，感知对向车流并等待安全转弯时机。在$t=21.0s$时(图\ref{fig:cross_sub_b})，DACER-F策略精准识别出车流中的安全间隙，开始加速并启动左转机动。随后，在$t=25.0s$至$t=28.5s$期间(图\ref{fig:cross_sub_c}, 图\ref{fig:cross_sub_d})，智能体以平滑的控制高效地穿越了路口。最终，在$t=34.5s$时(图\ref{fig:cross_sub_e})，车辆安全汇入目标车道。
    
    % 综合这两个案例，DACER-F展示了其在处理复杂交互场景时，决策的安全性、实时性与控制的平滑性的能力，成功地在任务效率与驾驶安全性之间取得了平衡。
    
    Fig.~\ref{fig:crossroad_main} visualizes the complex interaction of the DACER-F algorithm with oncoming traffic at an intersection. At $t=17.5\;\text{s}$ (Fig.~\ref{fig:crossroad_main}a), the agent (red vehicle) arrives at the intersection, perceives the oncoming traffic, and waits for a safe opportunity to turn. At $t=21.0\;\text{s}$ (Fig.~\ref{fig:crossroad_main}b), the DACER-F policy accurately identifies a safe gap in the traffic flow, then begins to accelerate and initiate the left-turn maneuver. Subsequently, during the $t=25.0\;\text{s}$ to $t=28.5\;\text{s}$ period (Figs.~\ref{fig:crossroad_main}c, d), the agent efficiently crosses the intersection with smooth control. Finally, at $t=34.5\;\text{s}$ (Fig.~\ref{fig:crossroad_main}e), the vehicle safely merges into the target lane.
    \begin{figure}[t]
        \centering
        
        % 第一个子图 (a)
        \subfloat[Iteration time]{%
            \includegraphics[width=0.5\linewidth]{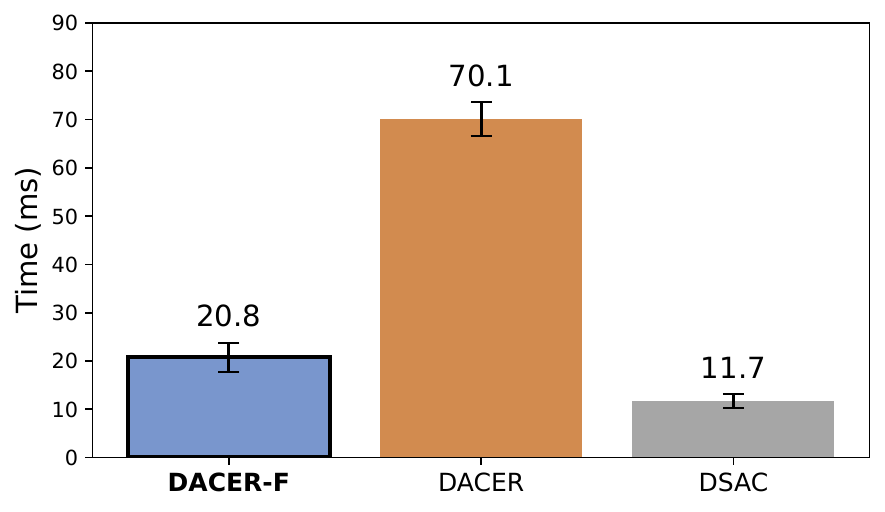}%
            \label{fig:algo_time}%
        }%
        \hfill % 在两个子图之间添加弹性水平间距
        % 第二个子图 (b)
        \subfloat[Inference time]{%
            \includegraphics[width=0.5\linewidth]{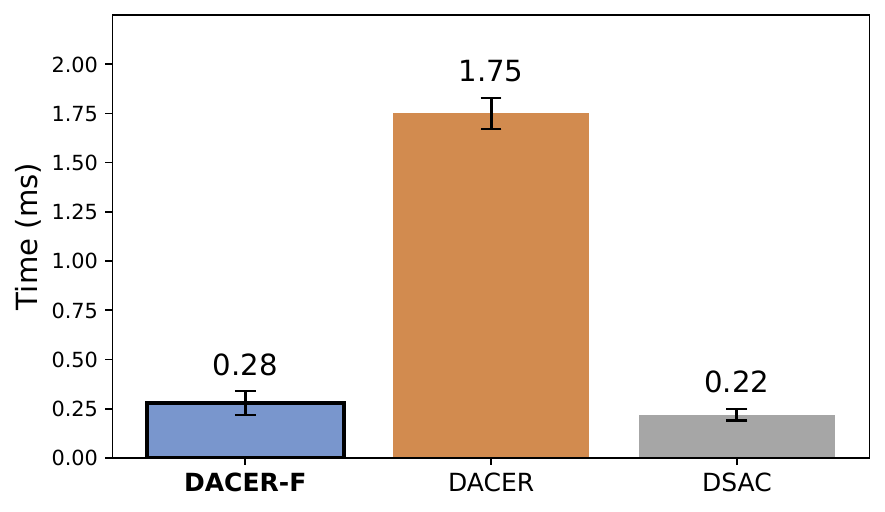}%
            \label{fig:infer_time}%
        }%
        
        \caption{Computational performance analysis. (a) The average iteration time reflects the training speed. (b) The inference time measures the execution latency of the policy.}
        \label{fig:performance_analysis}
    \end{figure}
    Combining these two cases, DACER-F demonstrates its capability for driving safety, real-time performance, and control smoothness when handling complex interaction scenarios, successfully achieving a balance between task efficiency and driving safety.
    
    \begin{figure*}[t]
        \centering
        \captionsetup[subfigure]{font=footnotesize}
    
        % 第一行：3张图，每张占约 0.32\linewidth
        \subfloat[\small Dog-run]{\includegraphics[width=0.32\linewidth]{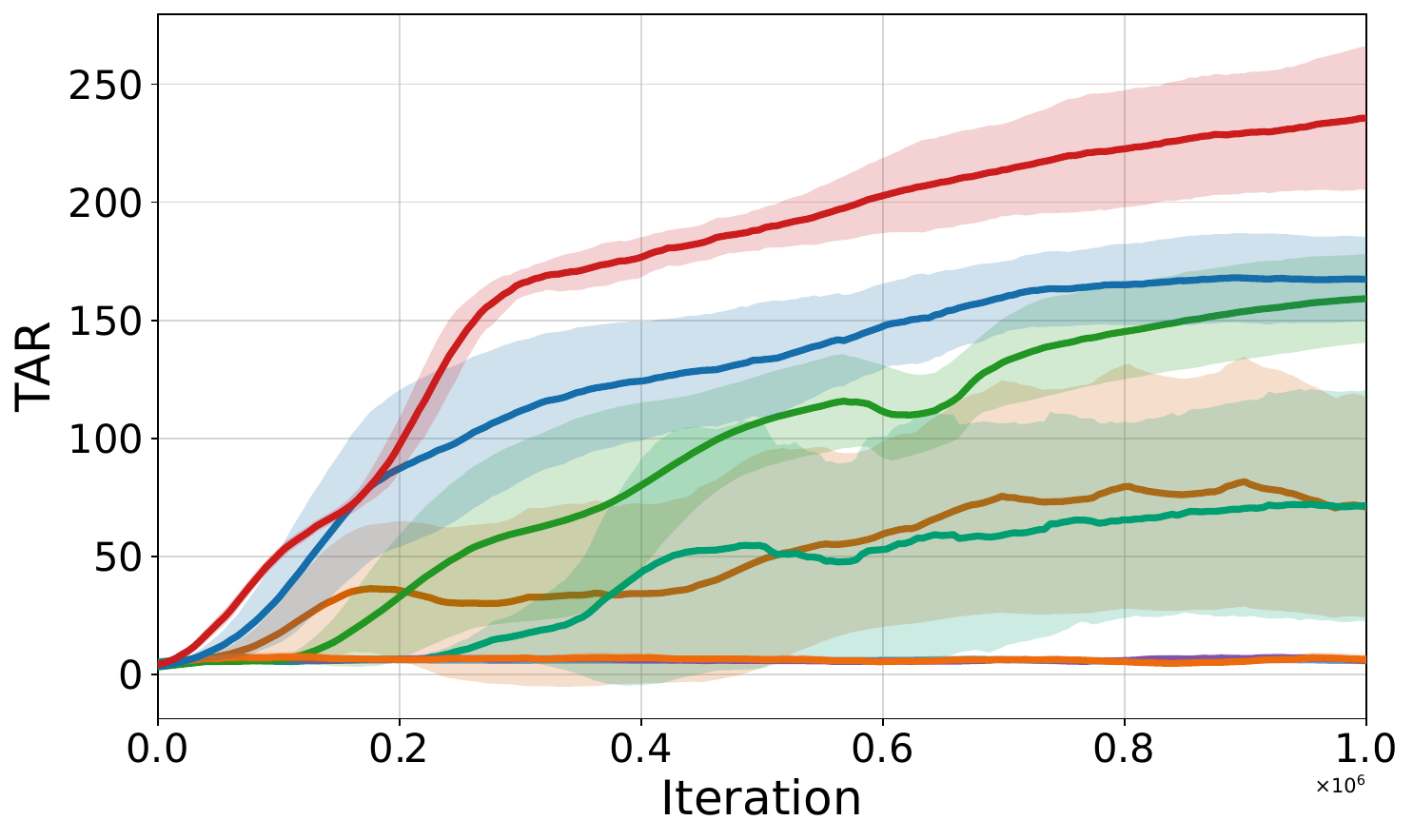}\label{fig:dog_run}} \hfill
        \subfloat[\small Dog-walk]{\includegraphics[width=0.32\linewidth]{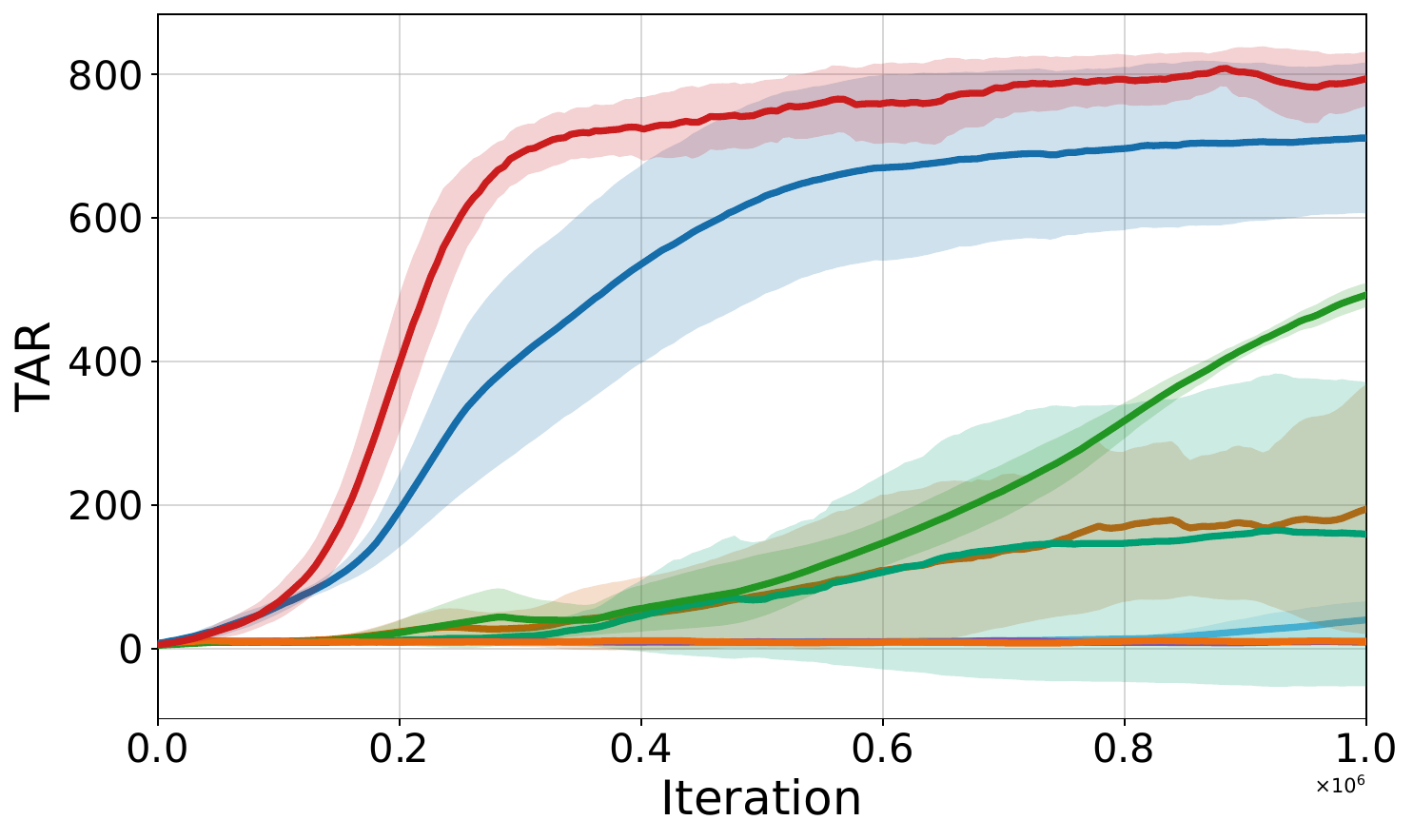}\label{fig:dog_walk}} \hfill
        \subfloat[\small Dog-stand]{\includegraphics[width=0.32\linewidth]{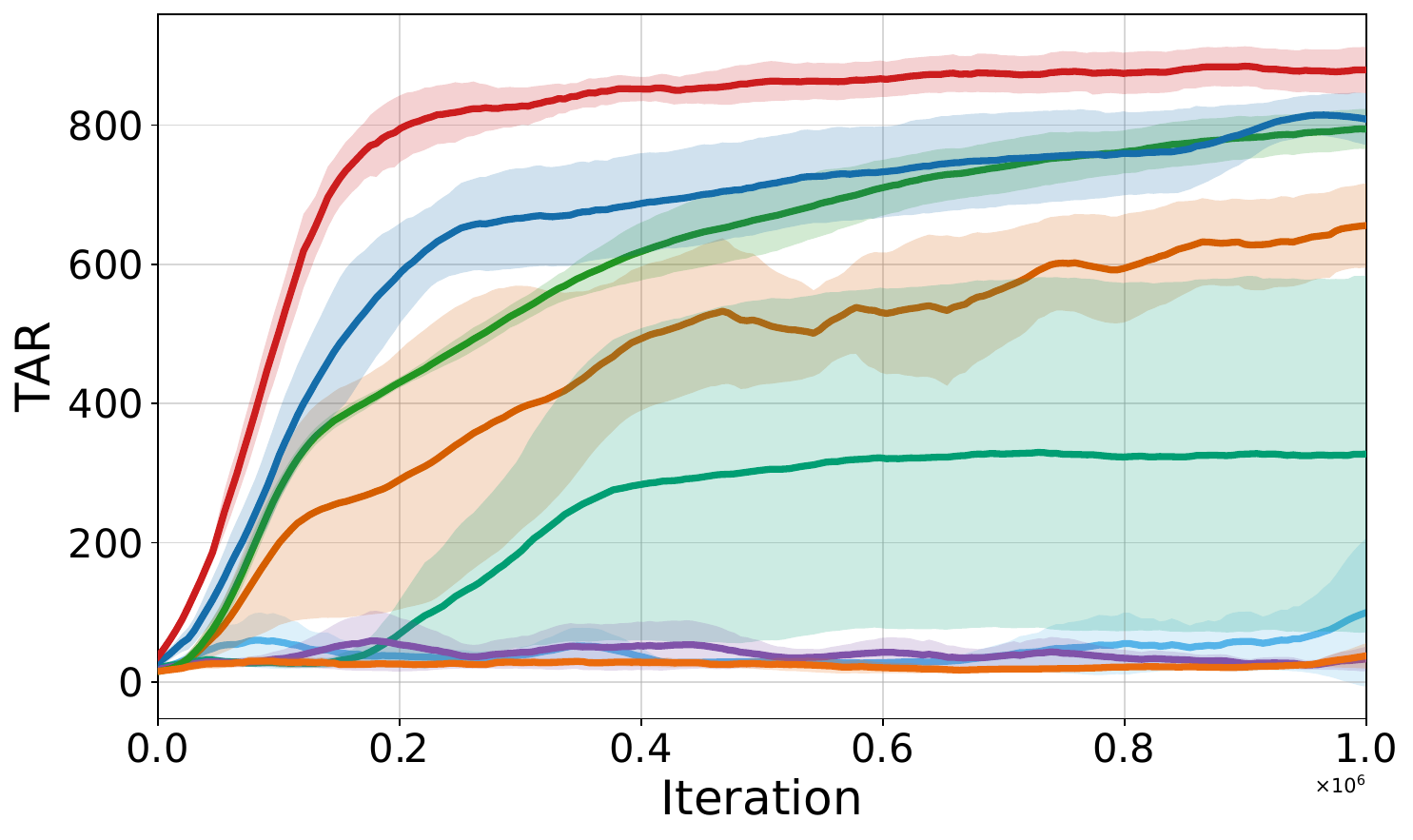}\label{fig:dog_stand}} \\ % 这里换行
        
        \vspace{1ex} % 添加第一行与第二行之间的垂直间距
        
        % 第二行：3张图，每张占约 0.32\linewidth
        \subfloat[\small Dog-trot]{\includegraphics[width=0.32\linewidth]{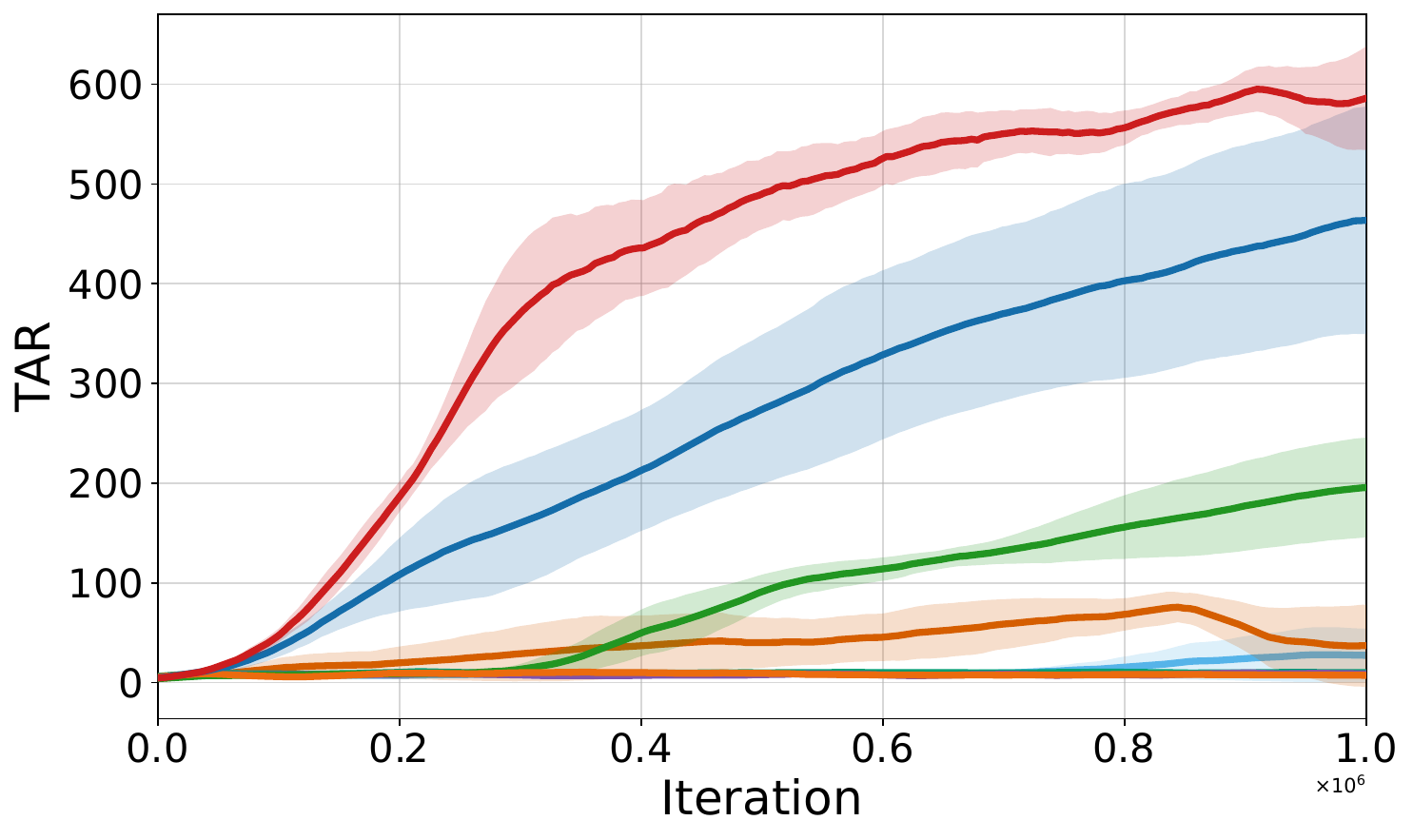}\label{fig:hm_run}} \hfill
        \subfloat[\small Humanoid-walk]{\includegraphics[width=0.32\linewidth]{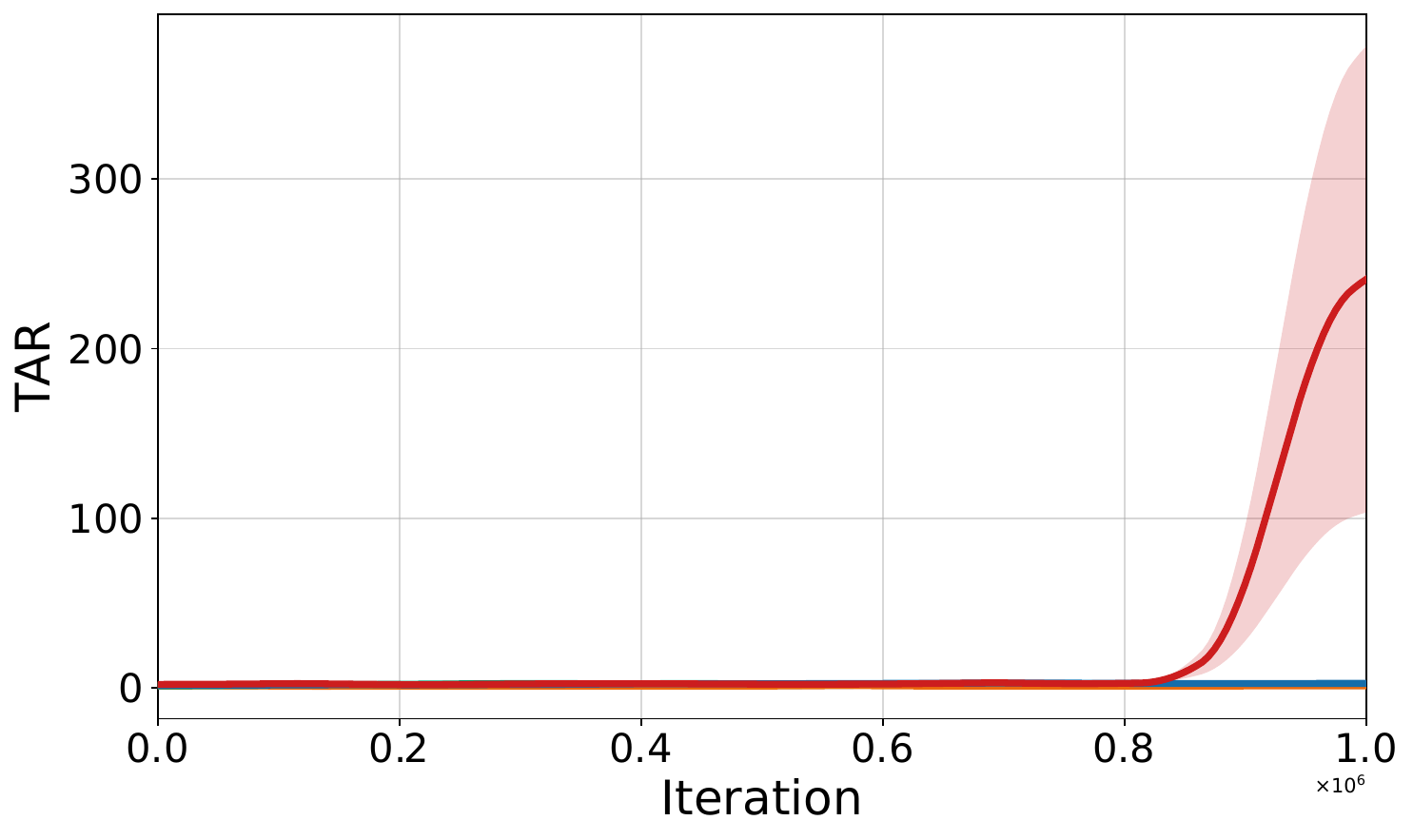}\label{fig:hm_walk}} \hfill
        \subfloat[\small Humanoid-stand]{\includegraphics[width=0.32\linewidth]{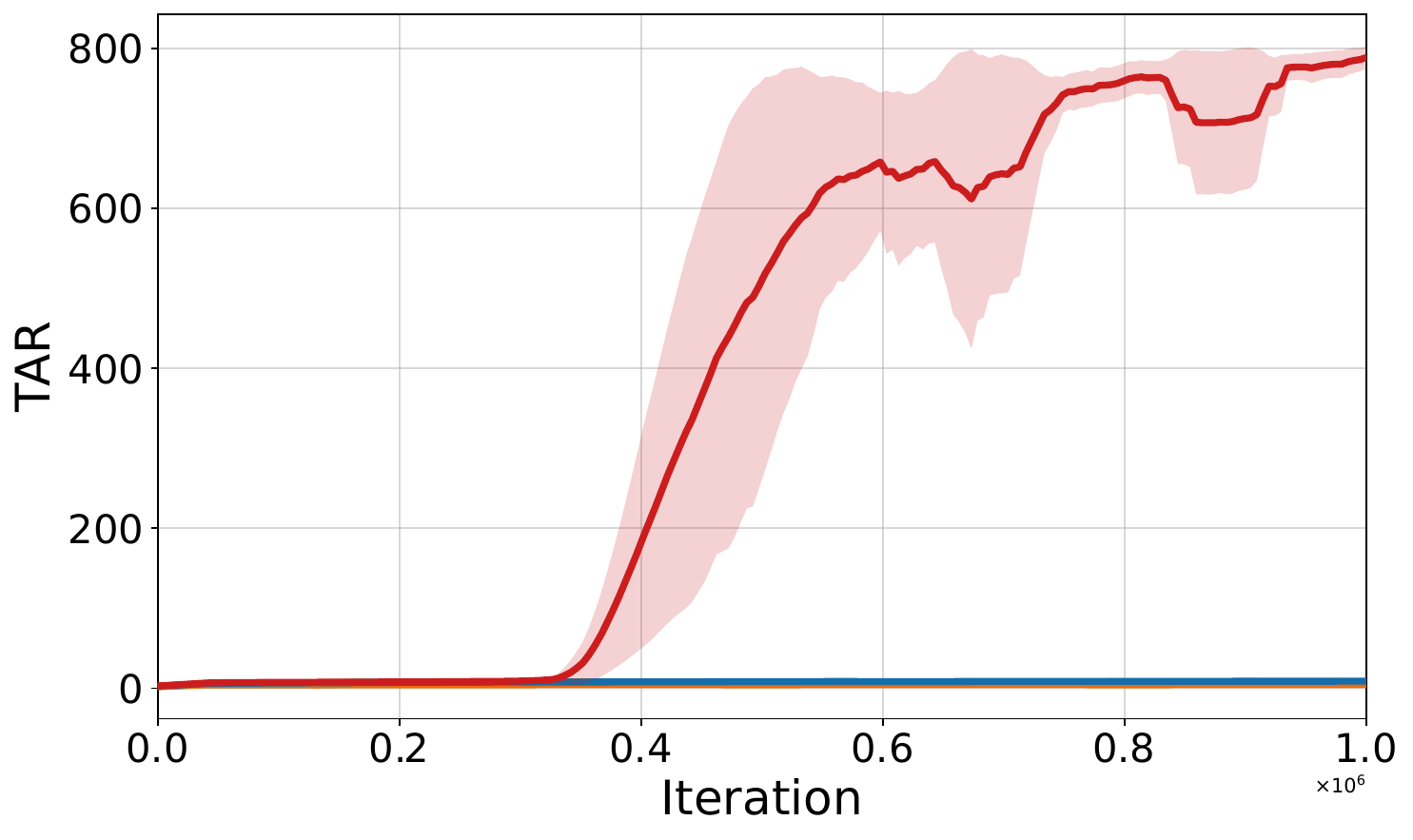}\label{fig:hm_stand}}
        
        \vspace{0.5ex} % 缩小图片与图例之间的距离
        
        % --- 底部中央图例 ---
        % 此时总图例的宽度可能也需要微调，0.8\linewidth 应该能横跨居中
        \includegraphics[width=0.8\linewidth]{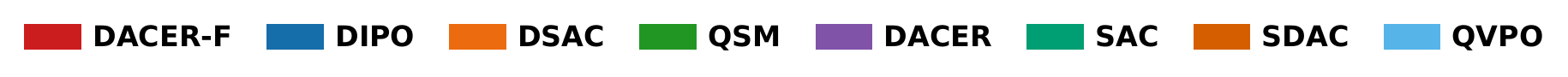} 
        
        \caption{Evaluating curves on benchmarks with 5 random seeds. The solid lines represent the mean, while the shaded regions indicate the confidence interval over five runs.}
        \label{fig:benchmark_curves}
    \end{figure*}
    
    \subsection{Time Efficiency}
    To evaluate the time efficiency of the algorithm, Fig.~\ref{fig:performance_analysis} compares the iteration time (training efficiency) and inference time (deployment efficiency) of DACER-F, DACER, and DSAC. Iteration time is defined as the average duration of a single training step, excluding the parallel environment interaction time, while inference time is the time required for the policy to generate a single action. All experiments were conducted using PyTorch, an Intel Core i9-12900K CPU, and an NVIDIA RTX 3090 Ti GPU.
    
    \begin{table*}[t]
        \caption{Average Return on DMControl benchmarks with 5 random seeds. Mean $\pm$ Std. \textbf{Bold} = best; Higher is better.}
        \centering
        \renewcommand{\arraystretch}{1.2} 
        % 调整列间距，确保 9 列在双栏页面中能放下
        \setlength{\tabcolsep}{5pt}      
        
        % 定义 9 列：l 代表环境列左对齐，8个 c 代表数据列居中对齐
        \begin{tabular}{lcccccccc}
        \toprule
        \textbf{Environment} & \textbf{DACER-F} & \textbf{DIPO} & \textbf{DSAC} & \textbf{QSM} & \textbf{DACER} & \textbf{SAC} & \textbf{SDAC} & \textbf{QVPO} \\
        \midrule
        % \multicolumn{9}{c}{\textbf{Humanoid Tasks}} \\
        % \midrule
        Humanoid-stand & \textbf{775.8 $\pm$ 12.7} & 8.7 $\pm$ 0.0 & 4.4 $\pm$ 0.4 & 8.1 $\pm$ 0.1 & 8.1 $\pm$ 0.4 & 6.9 $\pm$ 0.1 & 7.7 $\pm$ 0.0 & 7.6 $\pm$ 0.4 \\
        Humanoid-walk  & \textbf{226.5 $\pm$ 129.2}   & 2.5 $\pm$ 0.1 & 1.1 $\pm$ 0.2 & 1.9 $\pm$ 0.0 & 1.9 $\pm$ 0.8 & 2.3 $\pm$ 0.0 & 1.8 $\pm$ 0.0 & 1.6 $\pm$ 0.1 \\
        % \midrule
        % \multicolumn{9}{c}{\textbf{Dog Tasks}} \\
        % \midrule
        % 这里根据你的要求，将原来的 Humanoid-run 替换为了 Dog-trot
        Dog-trot       & \textbf{585.4 $\pm$ 28.8} & 449.9 $\pm$ 108.3 & 7.7 $\pm$ 3.2 & 187.5 $\pm$ 48.1 & 9.2 $\pm$ 1.4 & 10.0 $\pm$ 1.5 & 40.4 $\pm$ 34.2 & 27.3 $\pm$ 26.4 \\
        Dog-stand      & \textbf{879.2 $\pm$ 27.9} & 810.0 $\pm$ 27.5 & 25.1 $\pm$ 3.4 & 787.7 $\pm$ 28.2 & 25.6 $\pm$ 6.9 & 325.9 $\pm$ 253.8 & 639.0 $\pm$ 60.6 & 71.5 $\pm$ 61.3 \\
        Dog-walk       & \textbf{788.7 $\pm$ 37.4} & 707.4 $\pm$ 105.4 & 10.5 $\pm$ 2.0 & 458.3 $\pm$ 13.6 & 9.0 $\pm$ 0.7 & 162.4 $\pm$ 215.8 & 179.2 $\pm$ 138.1 & 31.1 $\pm$ 23.8 \\
        Dog-run        & \textbf{232.6 $\pm$ 27.3} & 167.3 $\pm$ 18.1 & 6.7 $\pm$ 1.4 & 157.0 $\pm$ 19.5 & 6.5 $\pm$ 0.5 & 71.5 $\pm$ 48.6 & 75.5 $\pm$ 48.4 & 6.2 $\pm$ 0.7 \\
        \bottomrule
        \end{tabular}
        \label{tab:rl_results_updated}
    \end{table*}
    
    % For training efficiency (Fig.~\ref{fig:algo_time}), our method (20.8 ms) significantly outperforms the other generative model, DACER (70.1 ms). Its training speed is 3.37 times faster (i.e., a 70.3\% reduction in training time). This is primarily attributed to the flow-matching architecture having a simpler and more efficient single-step training objective compared to the diffusion architecture. Although the iteration time of our method is higher than the lightweight MLP-based DSAC (11.7 ms) due to its more complex model structure, the comparison with DACER clearly demonstrates the training efficiency advantage of our method as a generative policy.
    
    For training efficiency (Fig.~\ref{fig:algo_time}), our method (20.8 ms) accelerates training by 3.37 times over DACER (70.1 ms) due to flow-matching's simpler single-step objective. Although naturally slower than the MLP-based DSAC (11.7 ms), our approach demonstrates exceptional training efficiency for a generative policy.

    In terms of inference efficiency (Fig.~\ref{fig:infer_time}), which is crucial for real-time control, DACER-F (0.28 ms) is 6.25 times faster than DACER (1.75 ms), yielding an 84.0\% time reduction. This acceleration results from replacing multi-step diffusion sampling with single-step generation. Notably, our 0.28 ms latency is highly competitive with the lightweight DSAC (0.22 ms), as both require only a single forward pass. Thus, DACER-F fully meets real-time standards, achieving MLP-level efficiency while preserving the powerful modeling capabilities of a generative policy.
    
    In conclusion, our method maintains powerful model expressiveness while achieving efficient training and extremely low inference latency, striking an excellent trade-off between model performance and practical deployment efficiency.
    
    \subsection{Scalability Analysis}

    To validate DACER-F as a general-purpose RL algorithm beyond the autonomous driving domain, we further benchmark it on standard continuous-control tasks. Specifically, we adopt DMC, a widely used continuous-control RL benchmark, and evaluate DACER-F on six challenging locomotion tasks. These tasks require high-dimensional control and precise balance/coordination, including agents such as humanoid (state/action dimensions: 67/24) and dog (state/action dimensions: 223/38). We compare DACER-F with seven representative online RL baselines, including SAC \cite{haarnoja2018soft} and DSAC \cite{DBLP:journals/pami/DuanWXGLLZCL25}, as well as expressive generative-policy methods: DIPO \cite{yang2023policy}, QSM \cite{DBLP:conf/icml/PsenkaEA024}, DACER \cite{wang2024diffusion}, SDAC \cite{DBLP:conf/icml/MaCW0025}, and QVPO \cite{ding2024diffusion}.

    The curves of the baseline experiment are illustrated in Fig. \ref{fig:benchmark_curves}. And the detailed numerical comparisons across 6 high-dimensional locomotion tasks are summarized in
    Table \ref{tab:rl_results_updated}. DACER-F consistently delivers the best TAR across all six tasks, demonstrating remarkable stability and effectiveness. Experimental results show that baselines such as DSAC and DACER exhibit near-zero performance on complex tasks like humanoid-stand. We hypothesize that DSAC’s degradation is related to the sparse-reward nature of DMC locomotion tasks, under which distributional Q estimation is more prone to pessimistic value predictions, thereby weakening the learning signal for policy improvement. In contrast, methods based on vanilla diffusion policies (e.g., DACER) can be more susceptible to unstable learning dynamics in online RL. Without an efficient guidance mechanism like our Langevin-guided dynamic targets, policy optimization may become difficult in high-dimensional state–action spaces due to the resulting complex optimization landscape.

    Notably, in the humanoid-stand task, DACER-F attains an average TAR of 775.8, substantially exceeding DACER (8.1) and SAC (6.9). In the humanoid-walk task, DACER-F achieves 226.5, again outperforming all baselines by a large margin, including DIPO (2.5) and SAC (2.3). This underscores the effectiveness of Langevin-guided flow matching in navigating complex energy distributions that traditional Gaussian or vanilla diffusion policies struggle to capture. In the dog-run task, DACER-F reaches 232.6, surpassing the strong DIPO baseline (167.3) by 39.0\%. Overall, DACER-F demonstrates strong performance on challenging locomotion tasks while preserving the real-time inference advantage of single-step flow matching.

    \section{Conclusion}
    % 在本文中，我们提出了DACER-F在动作梯度优化下的表达策略，旨在解决自动驾驶中强化学习多模态的挑战。我们使用flow-matching作为策略表达，并通过使用动作梯度优化的动作，能够利用更好的目标分布。在多车道和十字路口的大量模拟验证了所提出算法的有效性，展示了其在实际自动驾驶应用中的潜力。在未来的工作中，我们计划在更具挑战性的场景中评估算法的鲁棒性，以进一步评估其泛化能力。
    
    % 在本文中，我们提出了DACER-F，一种旨在解决自动驾驶中实时决策挑战的新型生成式策略。为解决现有生成式策略（如扩散模型）推理延迟高的关键问题，我们创新性地采用流匹配（Flow-matching）模型作为策略表达，以实现高效的单步采样。针对在线强化学习中缺乏“模仿目标”的核心难题，我们提出了一种基于朗之万动力学（Langevin Dynamics）的引导机制。该方法将Q函数视为一个隐式能量模型，通过采样生成一个高Q值且具探索性的动态目标分布，从而为流策略提供了稳定且有效的训练信号。在多车道和十字路口场景中的大量仿真验证了DACER-F的卓越性能。实验结果清晰地表明，我们的算法在总奖励、到达率和碰撞率等关键指标上显著优于DSAC和DACER等SOTA基线。更重要的是，DACER-F在保持强大性能的同时，实现了1.4ms的极低推理延迟，证明了其作为高性能、可实时部署的自动驾驶决策方案的巨大潜力。未来的工作将集中于在更复杂、更具挑战性的场景中评估算法的鲁棒性与泛化能力。

    In this paper, we propose DACER-F, a low-latency generative policy for real-time autonomous driving decision-making and control. We address two critical challenges: (i) the high inference latency of existing generative policies and (ii) the lack of reliable target distributions in online RL. To reduce inference latency, we adopt flow matching as the policy representation, enabling efficient single-step sampling. To stabilize learning, we introduce a Langevin dynamics guidance mechanism that leverages the Q-function as an implicit energy model to construct dynamic target distributions with improved quality. 
    Extensive driving simulations demonstrate the effectiveness of DACER-F, delivering strong performance across key metrics while maintaining an ultra-low inference latency of 0.28 ms. 
    Empirical results on DMC indicate that DACER-F generalizes beyond driving, reaching 775.8 on humanoid-stand and consistently outperforming competitive baselines.
    Future work will focus on robustness under more challenging and diverse conditions.
    
    % \newpage
    \bibliographystyle{IEEEtran}
    \bibliography{myReference} 
	
\end{document}